\documentclass[11pt]{article}

\usepackage[final]{acl}

\usepackage{times}
\usepackage{latexsym}

\usepackage[T1]{fontenc}
\usepackage[utf8]{inputenc}

\usepackage{microtype}

\usepackage{inconsolata}

\usepackage{amssymb}
\usepackage{amsmath}
\usepackage{graphicx}
\usepackage{tcolorbox}
\tcbuselibrary{listings, skins, breakable}
\usepackage{booktabs}
\usepackage{tabularx}
\usepackage{multirow}
\usepackage[table]{xcolor}
\usepackage{enumitem}
\usepackage{mdframed}
\usepackage{tikz}
\usetikzlibrary{positioning, shapes.geometric}
\usepackage{subcaption}

\definecolor{successcolor}{RGB}{0,120,60}
\definecolor{failcolor}{RGB}{180,30,30}
\definecolor{softblue}{RGB}{210,230,245}
\definecolor{lightblue}{RGB}{53,134,239}
\definecolor{lightgreen}{RGB}{106,168,79}
\definecolor{lightpurple}{RGB}{154,139,202}

\title{\textsc{FalsifyBench}: \\ 
Evaluating Inductive Reasoning in LLMs with Rule Discovery Games
}

\author{
 \textbf{Leonardo  Bertolazzi\textsuperscript{1}},
 \textbf{Katya Tentori\textsuperscript{1}},
 \textbf{Raffaella Bernardi\textsuperscript{2}}
\\
\\
 \textsuperscript{1}University of Trento,
 \textsuperscript{2}Free University of Bozen-Bolzano,
\\
 \small{
   \textbf{Correspondence:} \href{mailto:email@domain}{leonardo.bertolazzi@unitn.it}
 }
}

\begin{document}
\maketitle

\begin{abstract}
Large language models (LLMs) are increasingly deployed as autonomous agents in scientific tasks. Yet whether these systems can effectively engage in forms of inductive reasoning relevant to scientific discovery remains an open question. In this work, we introduce \textsc{FalsifyBench}, an evaluation framework for hypothesis-driven reasoning inspired by the classic Wason 2-4-6 task, in which agents must discover hidden semantic properties by iteratively proposing examples and receiving feedback. 
This task captures key elements of scientific reasoning: hypothesis generation, evidence gathering, and belief revision in response to both confirming and disconfirming evidence. Our evaluation of 12 LLMs across model families and scales shows that reasoning models are generally stronger scientific reasoners than instruction-tuned models, although no model comes close to optimal performance. The primary driver of success is the capacity for negative testing: models that actively seek to falsify their hypotheses consistently outperform those that primarily seek confirmation. Moreover, a fine-grained turn-level analysis, neglected in previous work, reveals that failure is tied to identifiable patterns in how models navigate the hypothesis space.
\end{abstract}

\section{Introduction}
\label{sec:intro}

Large language models (LLMs) are becoming increasingly capable across domains associated with scientific work, including literature review~\citep{mialon2024gaia, wei2025browsecomp}, software engineering and programming~\citep{jimenez2024swebench, jain2025livecodebench}, mathematics~\citep{mirzadeh2025gsmsymbolic, glazer2025frontiermath}, and expert-level knowledge~\citep{wang2024mmlupro, rein2024gpqa, phan2025lastexam}. These advances have motivated a growing body of work on autonomous agents capable of performing scientific research in an end-to-end manner~\citep{lu2024aiscientistfullyautomated, gottweis2025aicoscientist}. As these systems become increasingly integrated into scientific workflows, a fundamental question arises: can they carry out the key forms of inductive reasoning required for scientific research? Central among these is the ability to generate, evaluate, and revise hypotheses in light of new evidence. A critical component of this process is \emph{falsification}, whereby a hypothesis is rejected in favor of an alternative when it is contradicted by the evidence.

\begin{table}[!b]
\centering
\tiny
\begin{tabular}{lll}
\toprule
\textbf{Relation} & \textbf{Falsification of $H$} & \textbf{Strategy}\\
\midrule
$H = R$ & None & None \\
\rowcolor{softblue}
$H \subset R$ & $x \notin H \land x \in R$ & $-$ \\
$H \supset R$ & $x \in H  \land x \notin R$ & $+$ \\
Partial overlap & $(x \in H \land x \notin R) \lor (x \notin H \land x \in R)$ & $+/-$\\
Disjoint & $x \in H \lor x \in R$ & $+/-$\\
\bottomrule
\end{tabular}
\caption{Set relations between a possible hypothesis $H$ and the true rule $R$, where $x$ denotes a test triple. Each relation has specific \emph{falsification} cases that force the rejection of $H$ in favor of a new hypothesis. The highlighted case ($H\subset R$) corresponds to the Wason 2-4-6 task and the games in \textsc{FalsifyBench}: here, only triples outside $H$ (negative tests, $-$) can reveal that $R$ is broader than $H$, whereas for other relations triples inside $H$ (positive tests, $+$) may falsify H.}
\label{tab:conclusive}
\end{table}

One of the best-known and most widely used paradigms for studying this process in humans is the Wason 2-4-6 task~\citep{wason1960failure}. Participants are shown a triple of numbers (e.g., ``2, 4, 6'') and asked to discover the hidden rule underlying it, denoted here as $R$. They do so by proposing new triples and receiving feedback from the experimenter on whether each triple fits the rule. Once sufficiently confident that they have discovered $R$, participants state their hypothesis (denoted here as $H$). When the rule to be discovered is more general than the one suggested by the initial example (as in Wason’s original task, where $R$ is ``Any increasing sequence of three numbers''), success rates are low. This is because participants tend to form overly specific hypotheses (e.g., ``Even numbers increasing by two''). In such cases, generating triples consistent with specific hypotheses (e.g., ``8, 10, 12'') cannot reveal that those hypotheses are too narrow and falsify them. This behavior has been termed \emph{Confirmation bias} and has traditionally been interpreted as an irrational tendency that runs counter to scientific reasoning. 

Later work by \citet{klayman1987confirmation} has made a more nuanced distinction between  \emph{positive testing}, in which one seeks evidence expected to confirm the current hypothesis, and \emph{negative testing}, in which one seeks evidence expected to disconfirm it. Crucially, they showed that positive testing can also lead to the falsification of a hypothesis and that the effectiveness of these two strategies depends on the relation between $H$ and $R$.
Specifically, they showed that a positive test strategy can falsify a hypothesis $H$ when it is broader than the hidden rule $R$ ($H \supset R$), while it fails to falsify $H$ whenever this is a strict subset of $R$ ($H \subset R$). In the latter case, \emph{negative testing} is the only strategy that can conclusively falsify $H$. Table~\ref{tab:conclusive} provides a summary of all possible relations between $H$ and $R$, together with the type of test that can conclusively falsify $H$ in each case.\footnote{We include a detailed discussion of cases where $H$ can be conclusively falsified, with visual examples, in Appendix~\ref{app:falsification}.}

In this work, we introduce a generalized version of the Wason task in which LLMs must discover hidden semantic properties rather than numerical rules. In this semantic version, we exploit taxonomic relations to build games emphasizing the $H \subset R$ configuration, where negative testing is the optimal strategy and which corresponds to Wason’s original case.
This configuration is particularly interesting because it mirrors a recurring pattern in the history of science, where accepted theories later turn out to be special cases of broader ones (e.g., Newtonian mechanics as a limiting case of relativity). This parallel motivates our focus on the falsification and revision of overly narrow hypotheses through negative testing as a key component of scientific reasoning, and one worth testing in systems such as LLMs. At the same time, following \citet{klayman1987confirmation}, we do not assume that negative testing is universally superior. Instead, we examine how LLMs handle hypothesis testing across different relations between $H$ and $R$ and apply the most appropriate normative analysis to each case, as described below.

Our main contributions are as follows: (1) we introduce \textsc{FalsifyBench}\footnote{The code and data used for this paper are publicly available at \url{https://github.com/leobertolazzi/FalsifyBench.git}.}, a benchmark for hypothesis-driven reasoning grounded in semantic taxonomies, which operationalizes a key component of scientific reasoning: the ability to actively generate, and potentially falsify, a current hypothesis in order to control for it; (2) we evaluate 12 LLMs across multiple model families and scales, showing that reasoning models generally behave as stronger scientific thinkers than instruction-tuned models in our framework, although no model approaches optimal performance; and (3) we run a fine-grained turn-level analysis, made possible by our evaluation setting, showing that failures reflect identifiable patterns in how models navigate the hypothesis space.

\begin{figure*}[t]
\centering
\includegraphics[width=0.98\textwidth]{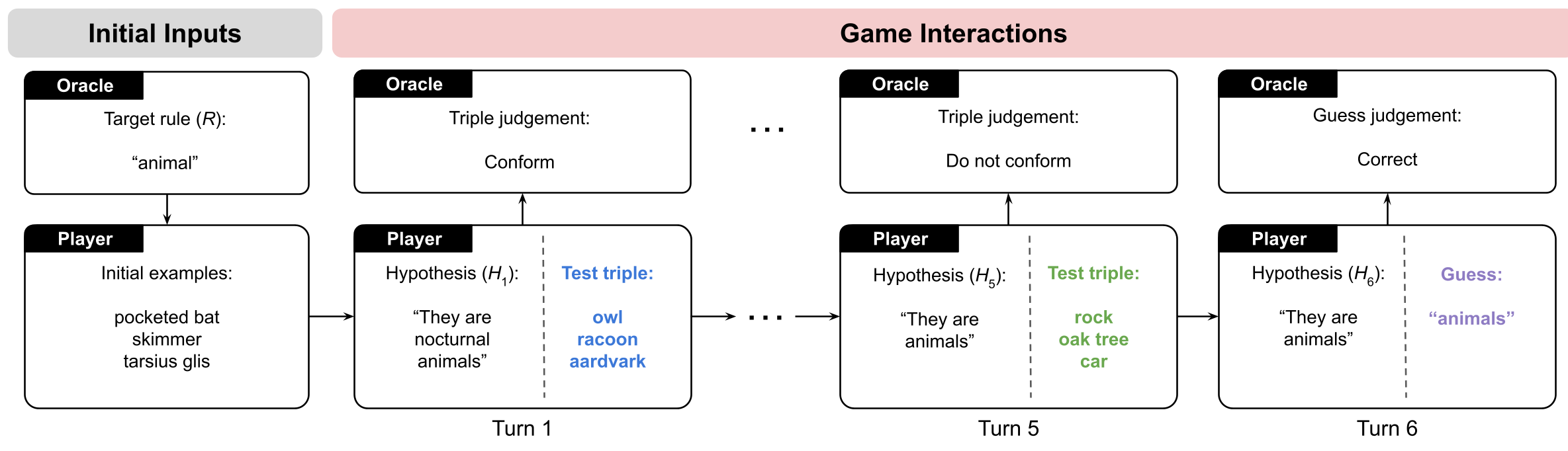}
\caption{ \textsc{FalsifyBench} game structure. The player is given three items (e.g., pocketed bat, skimmer, and 
tarsius glis) and must identify their shared hidden target rule (e.g., ``they are animals'') by proposing additional test triples and receiving feedback from the oracle (``Conform'' vs. ``Do not conform''). Before proposing a test triple, the player makes explicit the hypothesis about the hidden rule being tested. In \textcolor{lightblue}{\textbf{positive tests}}, the proposed triples are expected to satisfy this hypothesis, whereas in \textcolor{lightgreen}{\textbf{negative tests}}, they are expected to falsify it. The process continues until the player is sufficiently confident that they have identified the hidden rule and articulates it as a \textcolor{lightpurple}{\textbf{guess}}. If the stated hypothesis is correct, the game ends; otherwise, the player proceeds with a new test triple.}
\label{fig:game-example}
\end{figure*}

\section{Related Work}
\label{sec:related}

Inductive reasoning
is central to a wide range of human activities, including empirical scientific research, and has therefore become a key focus in evaluations of reasoning in LLMs.

Early studies have tested this ability in sequence-to-sequence models using synthetic benchmarks.
\citet{chollet2019arc} introduced the Abstraction and Reasoning Corpus (ARC), which tests models on grid transformation tasks, in which they must infer abstract transformations from a few example input-output pairs and learn to apply the correct rule to unseen grids. \citet{lake2018scan} take a similar approach, testing the capability to learn mappings from commands to action strings, and testing their application to novel, unseen compositions of sequences.

Recent work has increasingly focused on understanding how LLMs perform inductive reasoning beyond task performance. \citet{jin2026evaluating} showed that chain-of-thought reasoning \citep{wei2022cot} does not universally improve induction and can even reduce performance. \citet{li2025mirage} introduced a framework for evaluating and explaining inductive reasoning in LLMs by disentangling induction and deduction steps. \citet{louapre2026gameofscience} examined how LLMs behave in an inductive game, explicitly mapping models' strategies to human-like traits that influence performance, and connecting their setting to scientific discovery.

Within this line of research, our work is closely related to studies that also draw inspiration from Wason’s 2-4-6 task. \citet{banatt2024wilt} introduce WILT, a multi-turn inductive logic benchmark where hidden rules are boolean functions involving three variables; \citet{jhaveri2026falsify} use rules about number properties and explicitly study Confirmation bias in LLMs, applying interventions studied in humans to LLMs to encourage falsificatory behavior. Neither study explicitly takes into account the normative framework introduced by \citet{klayman1987confirmation}: 
they neither sample triples from a rule that is a strict subset of the general target rule nor track how the hypotheses formed by models at each turn relate to the target rule. In our experiment, we instead address this limitation by sampling triples of terms from nodes that are strict taxonomic descendants of $R$, thereby making negative testing the better strategy for identifying it.

\begin{figure*}[t]
    \centering
    % Column headers
    \begin{minipage}[b]{0.48\textwidth}
        \centering
        {\sffamily\large\textbf{Player}}
    \end{minipage}
    \hfill
    \begin{minipage}[b]{0.48\textwidth}
        \centering
        {\sffamily\large\textbf{Oracle}}
    \end{minipage}
    \\[4pt]
    % Top row
    \begin{minipage}[b]{0.48\textwidth}
        \includegraphics[width=\linewidth]{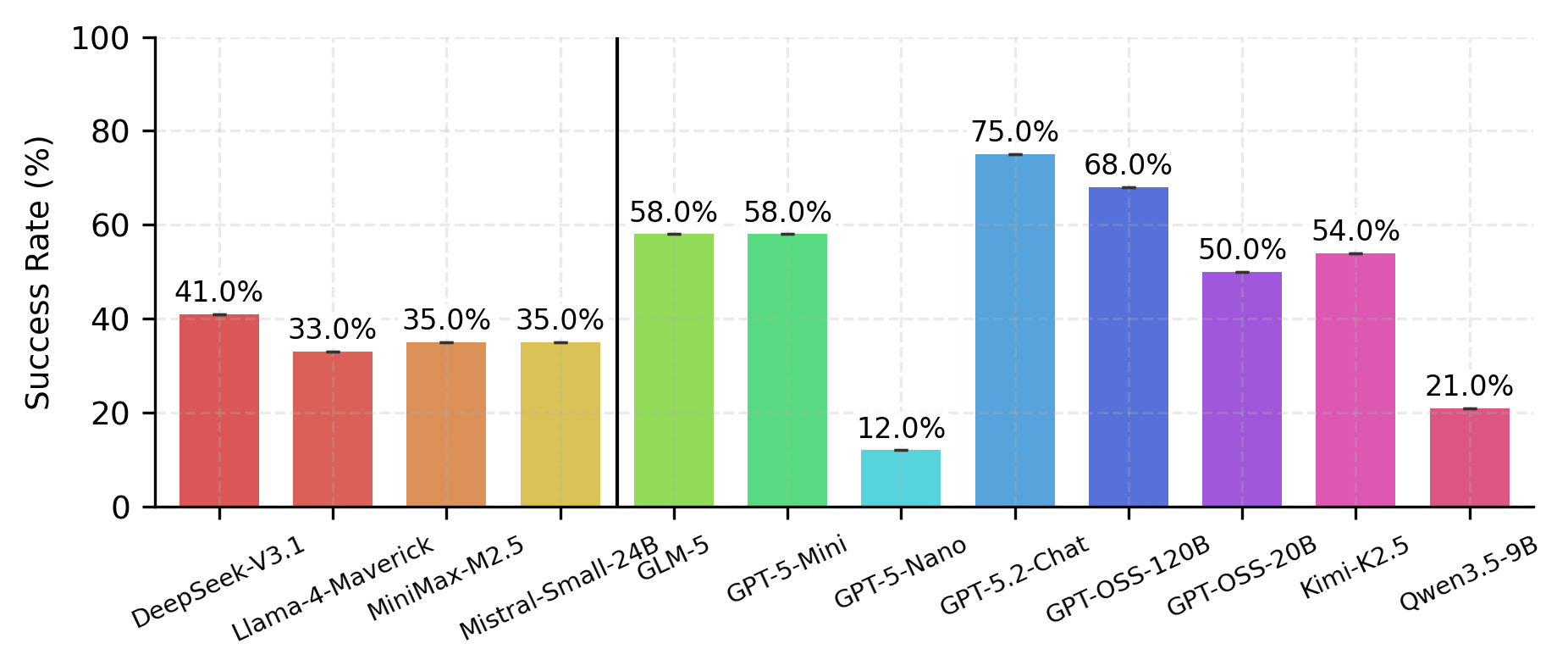}
    \end{minipage}
    \hfill
    \begin{minipage}[b]{0.48\textwidth}
        \includegraphics[width=\linewidth]{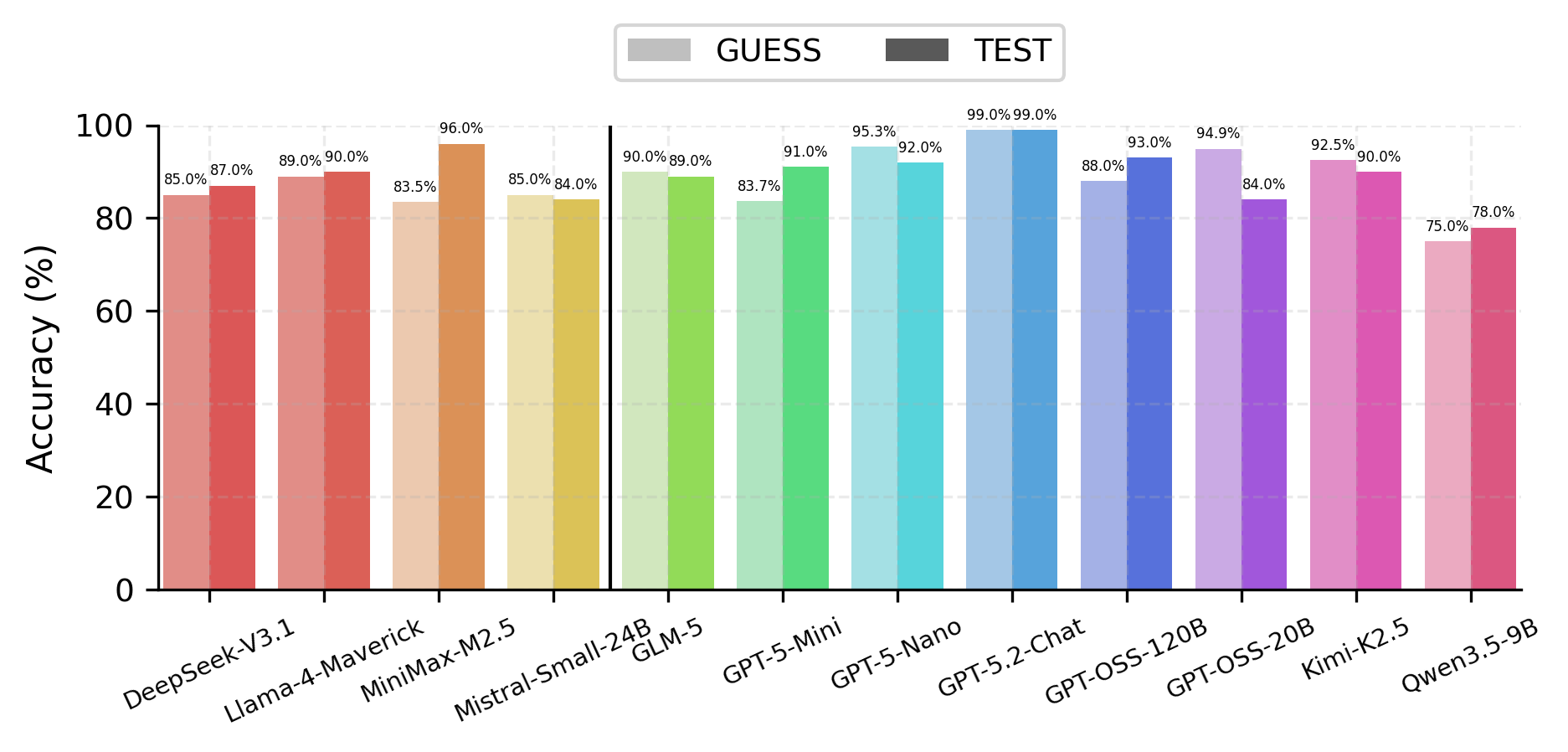}
    \end{minipage}
    \\[6pt]
    % Bottom row
    \begin{minipage}[b]{0.48\textwidth}
        \includegraphics[width=\linewidth]{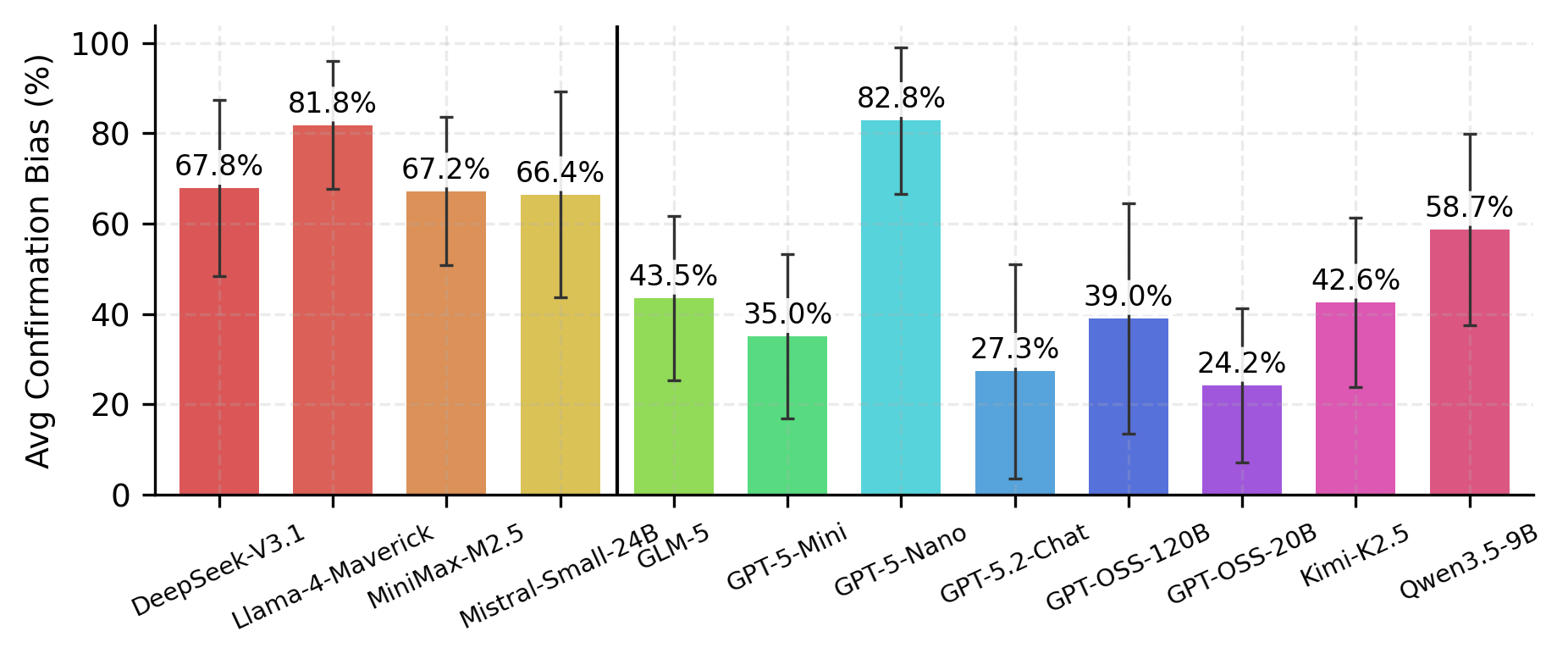}
    \end{minipage}
    \hfill
    \begin{minipage}[b]{0.48\textwidth}
        \includegraphics[width=\linewidth]{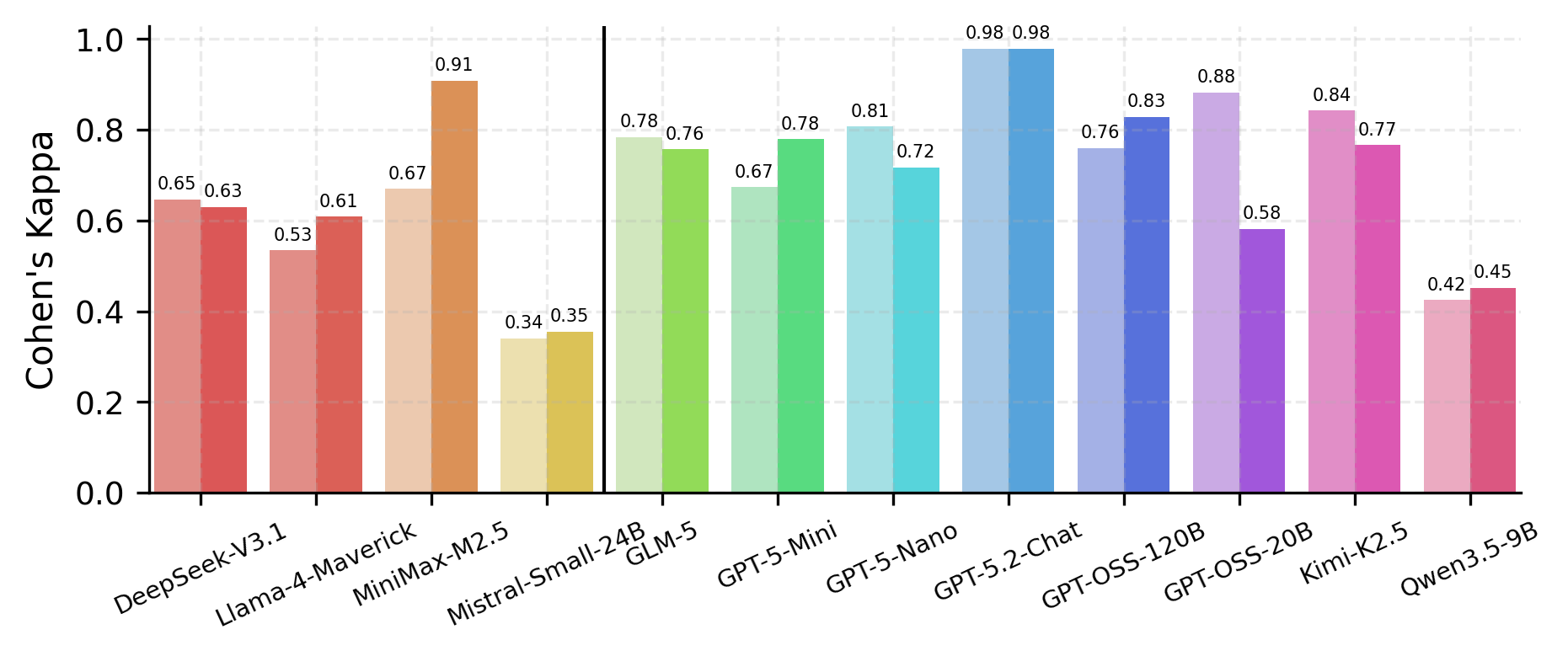}
    \end{minipage}
    \caption{Experimental results for 12 LLMs on \textsc{FalsifyBench}. Models to the left of the vertical black line are \textbf{instruction-tuned} variants, while models to the right are \textbf{reasoning} variants. \textbf{Top-left:} Success rate represents the percentage of correctly identified target rules. \textbf{Bottom-left:} Confirmation bias measures the proportion of queries that confirm rather than falsify the current hypothesis (lower is better). \textbf{Top-right:} Oracle accuracy against human annotations on \texttt{Test} and \texttt{Guess} turns. \textbf{Bottom-right:} Cohen's $\kappa$ agreement between oracle and human annotations on \texttt{Test} and \texttt{Guess} turns.}
    \label{fig:results_main}
\end{figure*}

\section{Method}

\subsection{Experiment Overview}
\label{sec:exp_setup}

In our experiment, each LLM served in two roles: as the oracle, which answered queries, and as the player, which attempted to discover the hidden rule. In direct analogy with Wason’s original experiment, the oracle provided the player with an initial triple of terms sampled from a subordinate semantic category (e.g., ``vertebrate''), which was itself included in a more general target category (e.g., ``animal''). The more general target category served as the target rule $R$. By iteratively proposing new examples and receiving feedback, the player accumulated evidence, testing its current hypothesis either positively or negatively, until it could identify $R$. An example of a real game is shown in Figure~\ref{fig:game-example}. More specifically, on each turn, the player proposed three items along with a hypothesis and a rationale for why those items were chosen. The oracle responded with the string ``Conform'' or ``Do not conform'' depending on whether the set of items satisfied the target rule $R$. When sufficiently confident, the player submitted a final answer, and the oracle responded with either ``Correct'' or ``Incorrect''. Thus, the player was prompted to use \texttt{Test} actions to gather information and refine its hypotheses, and to use \texttt{Guess} actions once enough information had been collected. When the player submitted an incorrect guess, it learned that its current hypothesis was wrong and was prompted to perform new actions to gather additional evidence until it was confident enough to make a new guess. A game ended when either the player submitted a correct guess or the maximum turn limit was reached (we set the maximum number of turns to 20). 

We chose semantic categories, rather than numerical sequences, as the domain of our experiment for two reasons: LLMs are very likely to have encountered the classic 2-4-6 task during pretraining, and semantic taxonomies allow us to define a rich set of candidate concepts with increasingly specific subconcepts. The categories were extracted from WordNet~\citep{miller1995wordnet}, which enables the construction of a large number of such games by selecting general concepts and sampling from their descendants. From this large space, we curated a set of 100 representative games spanning different taxonomic depths and semantic domains.

Each game was defined by two semantic categories: the first was the general category that served as the target hidden rule $R$, while the second was a sampling category, $S$, a descendant of $R$ from which the initial examples were drawn. For instance, if $R =$ ``animal'' and $S =$ ``vertebrate'', a possible set of items was $E =$ \{``pocketed bat'', ``skimmer'', ``tarsius glis''\}. The player observed these items and formed an initial hypothesis, $H_1$. Because the items were drawn from the node $S$, which was deep in the taxonomy, semantic categories close to that region tended to appear more plausible than very general ones, like $R$, so players typically began with a hypothesis that was more specific than the true rule.\footnote{This pattern mirrors Wason's findings, where participants exposed to the sequence ``2-4-6'' favored overly specific rules such as ``increasing even numbers'' instead of the true rule ``any increasing sequence''.}
%The player then iteratively tested its hypothesis by proposing triples and receiving feedback on whether or not they satisfied $R$. 
When $H$ was more specific than $R$, \emph{positive testing} always returned positive feedback. Making progress therefore required \emph{negative testing}: only by proposing items outside the current hypothesis could the player determine that $R$ was broader than hypothesized and revise the hypothesis upward toward it.

Overall, we selected 7 high-level WordNet categories as candidate target rules: ``animal'', ``artifact'', ``body part'', ``food'', ``location'', ``plant'', and ``worker''. For each, we identified all descendant subcategories and sampled triples from their leaves, yielding 17,905 total games. Since many WordNet subcategories involve low-frequency vocabulary, we manually filtered the subcategory pool and created a curated set of 100 games spanning 5 categories: ``animal'', ``artifact'', ``body part'', ``food'', and ``plant''. Full details on game construction and the complete list of selected sampling categories appear in Appendix~\ref{app:games}. The results we report in this paper are collected on this curated set of 100 games.

The two roles assumed by each LLM (i.e,  player and oracle) were instantiated through separate API calls. This allowed us to evaluate models not only on hypothesis-driven reasoning, but also on their ability to provide semantic judgments consistent with the taxonomic knowledge encoded in WordNet. We forced both agents to produce actions in JSON format to constrain the space of their possible outputs. We provide more details on the exact prompts and JSON schema used in our experiments in Appendix \ref{app:game_engine}.

\begin{figure*}[t]
    \centering
    \includegraphics[width=\linewidth]{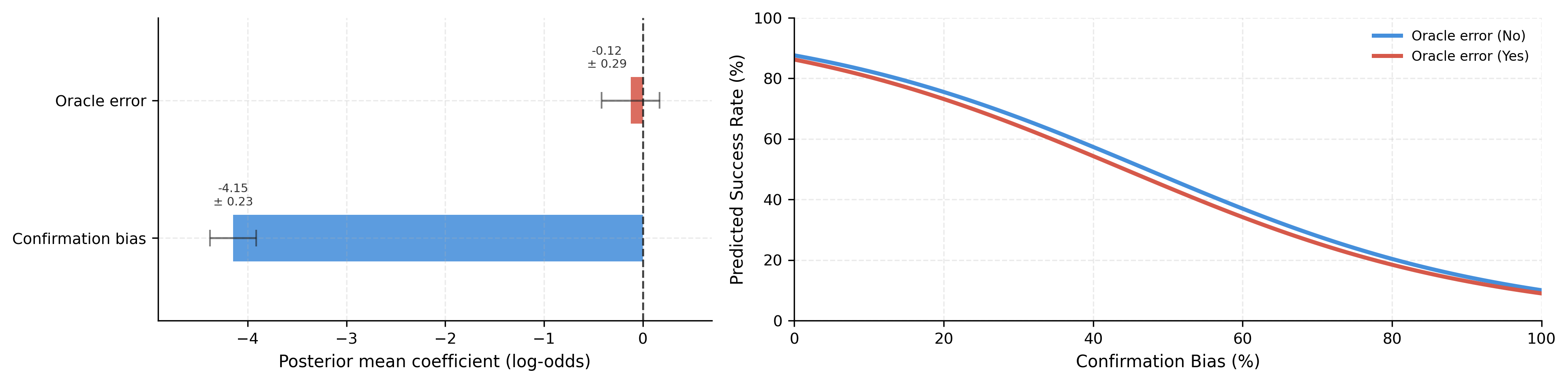}
    \caption{Results of the Bayesian mixed-effects logistic regression ($n = 1{,}200$ games). \textbf{Left:} Posterior means ($\pm 2$ posterior SD) for the two fixed-effect predictors. Confirmation bias is a strong negative predictor of success, while oracle error rate has a negligible effect. \textbf{Right:} Predicted probability of success as a function of Confirmation bias, shown separately for games in which the oracle made at least one error and games in which it did not. The near-identical curves indicate that oracle error rate is not the primary driver of player outcomes.}
    \label{fig:mixed_effects}
\end{figure*}

\subsection{Metrics}

We evaluate performance of LLMs on \textsc{FalsifyBench} using the following metrics.

\paragraph{Player Success Rate.} We computed the percentage of games in which the player correctly identified the target rule $R$. This served as the primary measure of task performance, indicating whether the model could converge on the correct hypothesis.

\paragraph{Player Confirmation Bias.}
We tracked the proportion of positive tests among all tests performed within a game to assess whether models instantiated a general positive testing strategy, as humans often do in the original Wason task. We then averaged this proportion across games and refer to it as Confirmation bias. By construction, in our experiment, positive tests were expected to be largely ineffective at falsifying the current hypothesis, making this metric a direct measure of the extent to which a model pursues an uninformative strategy.

\paragraph{Oracle Taxonomic Knowledge.}
Since each LLM served in both roles (i.e., as player and as oracle), we quantified the oracle's taxonomic knowledge to assess whether potentially inaccurate feedback could contribute to poor player performance. To this end, for each game and each model, we randomly sampled one triple from the \texttt{Test} turns and one hypothesis from the \texttt{Guess} turns.\footnote{With 100 games and 12 models, this yields 1200 \texttt{Test} instances and 1200 \texttt{Guess} instances.} A human annotator then labeled the oracle’s feedback for each triple and each guess according to whether it conformed to $R$. The annotator was given access to the NLTK library~\citep{bird2004nltk} and to search engines to ensure that annotations accurately reflected the taxonomic relations encoded in WordNet. We evaluated oracle performance by measuring its accuracy and agreement with the human annotations, using Cohen's $\kappa$.

\section{Performance on \textsc{FalsifyBench}}
\label{sec:results}

Our evaluation across 12 models revealed substantial variation in player and oracle performance, which we analyze below.

\paragraph{Player Performance and Confirmation Bias.}
Reasoning models generally achieved higher success rates than their instruction-tuned counterparts. The best instruction-tuned model, DeepSeek-V3.1, reached only 41\% success rate, while top reasoning models achieved 75\% (GPT-5.2-Chat) and 68\% (GPT-OSS-120B), with the remaining instruction-tuned models clustering between 33\% and 35\%. Notably, success rates remained far from saturated even for the best model, leaving substantial room for improvement. The advantage of reasoning models did not uniformly extend to the smallest ones: GPT-5-Nano (12\%) and Qwen3.5-9B (21\%) performed worse than all instruction-tuned models, suggesting that extended thinking alone was insufficient at smaller scales.
Additionally, we found a strong correlation between success rates and Confirmation bias: models that predominantly used positive testing consistently underperformed those that actively tested outside their current hypothesis (Spearman $\rho = -0.779$, $p = 0.003$). This pattern held across both model families and scales, with stronger reasoning models maintaining substantially lower Confirmation bias ratios (GPT-5-Mini: 39.0\%, GPT-5.2-Chat: 27.3\%, GPT-OSS-20B: 24.2\%) compared to instruction-tuned models, which clustered between 66\% and 82\%.

\paragraph{Oracle Quality.}
In terms of oracle performance with respect to human annotation, raw accuracy was moderately high across all models, with most LLMs clustering between 85\% and 95\%. Cohen's $\kappa$ revealed more meaningful variation, spanning from 0.34 (Mistral-Small-24B, \texttt{Guess} turns) to 0.98 (GPT-5.2-Chat, both conditions), indicating that some models provided less reliable taxonomic judgments than others. However, models with comparable oracle performance showed substantially different success rates: for instance, MiniMax-M2.5, GLM-5, and GPT-5-Nano achieved broadly similar oracle $\kappa$ scores (0.67--0.91, 0.78--0.76, and 0.81--0.72, respectively) yet succeeded on 35\%, 58\%, and 12\% of games respectively. This dissociation suggested that oracle quality was not sufficient for player success. We analyzed this further below.

\paragraph{Disentangling Player Reasoning from Oracle Quality.}
We aimed to better assess whether player strategy or oracle quality were determinant for the player's success. To this end, we fit a Bayesian mixed-effects logistic regression at the game level ($n = 1{,}200$), with Confirmation bias and oracle error rate as fixed-effect predictors and LLM identity as a random effect.\footnote{See Appendix~\ref{app:mixed-effects} for complete model details and results.} 
Confirmation bias was used in the model as a continuous predictor in $[0, 1]$, representing the proportion of positive tests within a game. Oracle error rate was modeled as a binary variable, taking the value 1 if the oracle's response disagreed with the human annotator on either the sampled \texttt{Test} or \texttt{Guess} turn for that game, and 0 otherwise.\footnote{Although an exhaustive human annotation of all turns across 1,200 games would enable us to calculate an exact continuous oracle error rate, we restricted our analysis to this binary estimate to keep the annotation effort manageable.}
As shown in Figure~\ref{fig:mixed_effects} (left), Confirmation bias was a strong negative predictor of success (posterior mean $= -4.15$, OR $= 0.016$, posterior SD $= 0.116$), while oracle error rate showed no credible independent effect (posterior mean $= -0.12$, OR $= 0.885$, posterior SD $= 0.147$), with a credible interval spanning zero. Moreover, Figure~\ref{fig:mixed_effects} (right) showed near-identical predicted success curves across oracle error conditions, suggesting that failures were driven primarily by the model’s own Confirmation bias rather than by the quality of the oracle's feedback. The residual between-model variance captured by the random effect ($\hat{\sigma}_u = 0.553$) indicates that LLM identity provides additional explanatory information beyond the two predictors. This reflects the factors that differentiate the overall quality of LLMs.

\begin{figure}[t!]
    \centering
    \includegraphics[width=\linewidth]{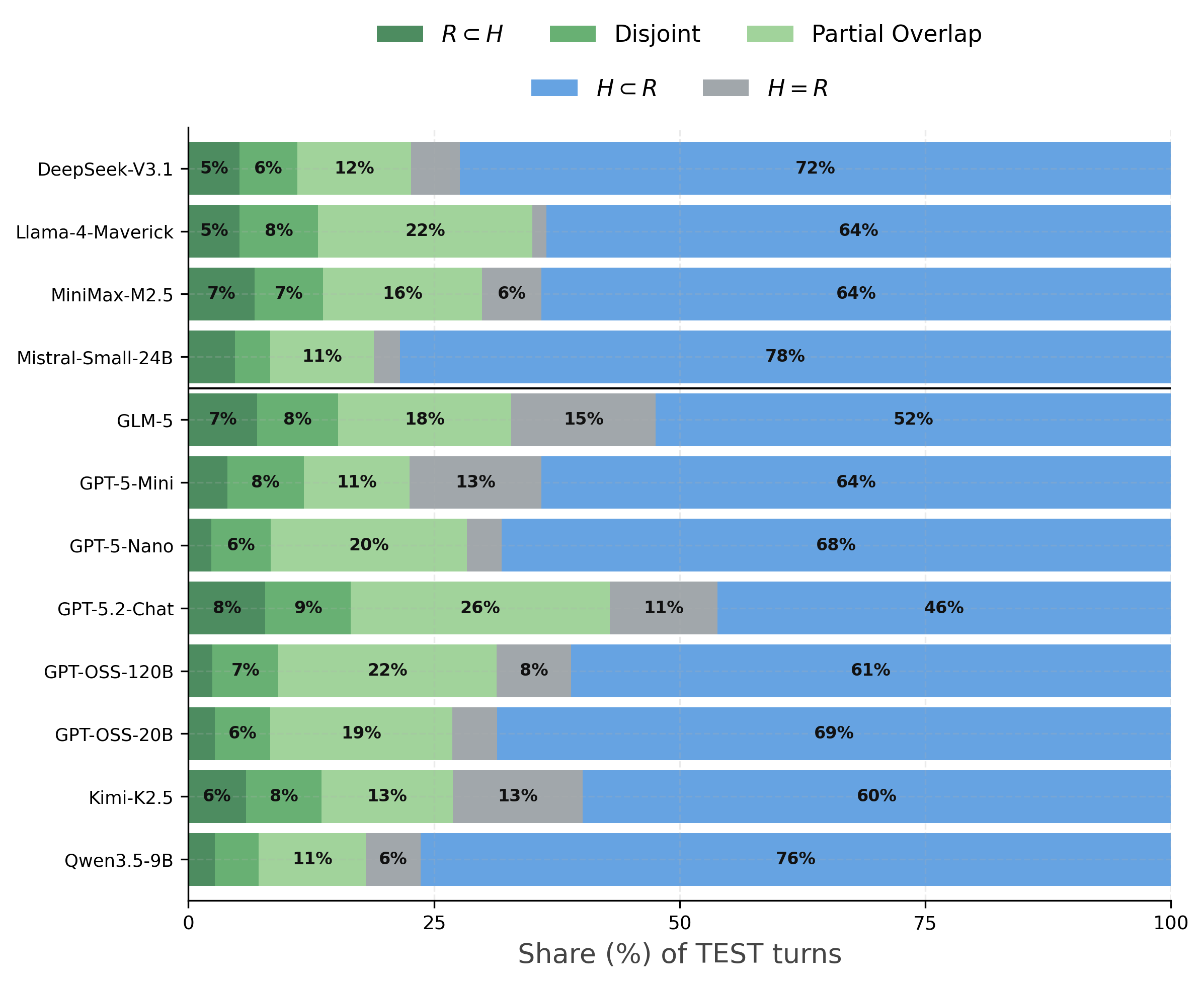}
    \caption{Distribution of relations between the current hypothesis $H$ and target $R$ across all \texttt{Test} turns, per model. The dominant configuration, $H \subset R$ (blue), is the only case in which positive testing is uninformative and negative testing is the only type of action that can result in a falsification of $H$. The green relations are ones where positive testing may also lead to a falsification of $H$. In the gray relation, $H = R$, no falsification is possible.}
    \label{fig:relation_distribution}
\end{figure}

\section{In-depth Turn Level Analysis}
\label{sec:conclusive_falsifications}

The Confirmation bias reported in Section~\ref{sec:results} measures the \emph{intent} behind a model's testing strategy, implicitly assuming that positive testing is a less optimal behavior than negative testing. In order to determine the effectiveness of a given strategy in rejecting a wrong hypothesis, it is necessary to understand the relationship between each current hypothesis $H$ and the target rule $R$ in games (see Table~\ref{tab:conclusive}). Indeed, it is possible that the player would form hypotheses for which negative testing is suboptimal and positive testing should instead be favored. For this reason, we employed one of the models tested on \textsc{FalsifyBench} as an offline LLM annotator, which annotates all \texttt{Test} turns of each game across all models with the set relation between $H$ and $R$. To balance speed and annotation cost, we selected GPT-5-Mini as annotator. The exact prompt used to perform this annotation can be seen in Appendix~\ref{app:game_engine}.

\paragraph{Hypothesis-Target Relations.}
Figure~\ref{fig:relation_distribution} shows the distribution of relations between $H$ and $R$ across all \texttt{Test} turns from all games for each model. The dominant configuration across all models is $H \subset R$ (Hypothesis in Target), consistent with the assumption we made when building \textsc{FalsifyBench}: since the initial items are drawn from a semantic category $S$ which is a descendant of $R$, models will form hypotheses that are more specific than the true target and revise them upward, yielding a sequence of $H \subset R$ turns. For almost all models, the relation $H \subset R$ was present in more than 50\% of all turns, with the exception of GPT-5.2-Chat, which exhibited a lower share of $H \subset R$ turns and a higher share of partial-overlap relations. Importantly, the $H \subset R$ configuration is the only one in which negative testing is the uniquely optimal strategy, and it remained the most common throughout games. In the cases where $H$ is partially overlapping with or disjoint from $R$, both positive and negative testing may lead to a falsification of $H$ (see Table~\ref{tab:conclusive}).

\begin{figure}[t!]
    \centering
    \includegraphics[width=\linewidth]{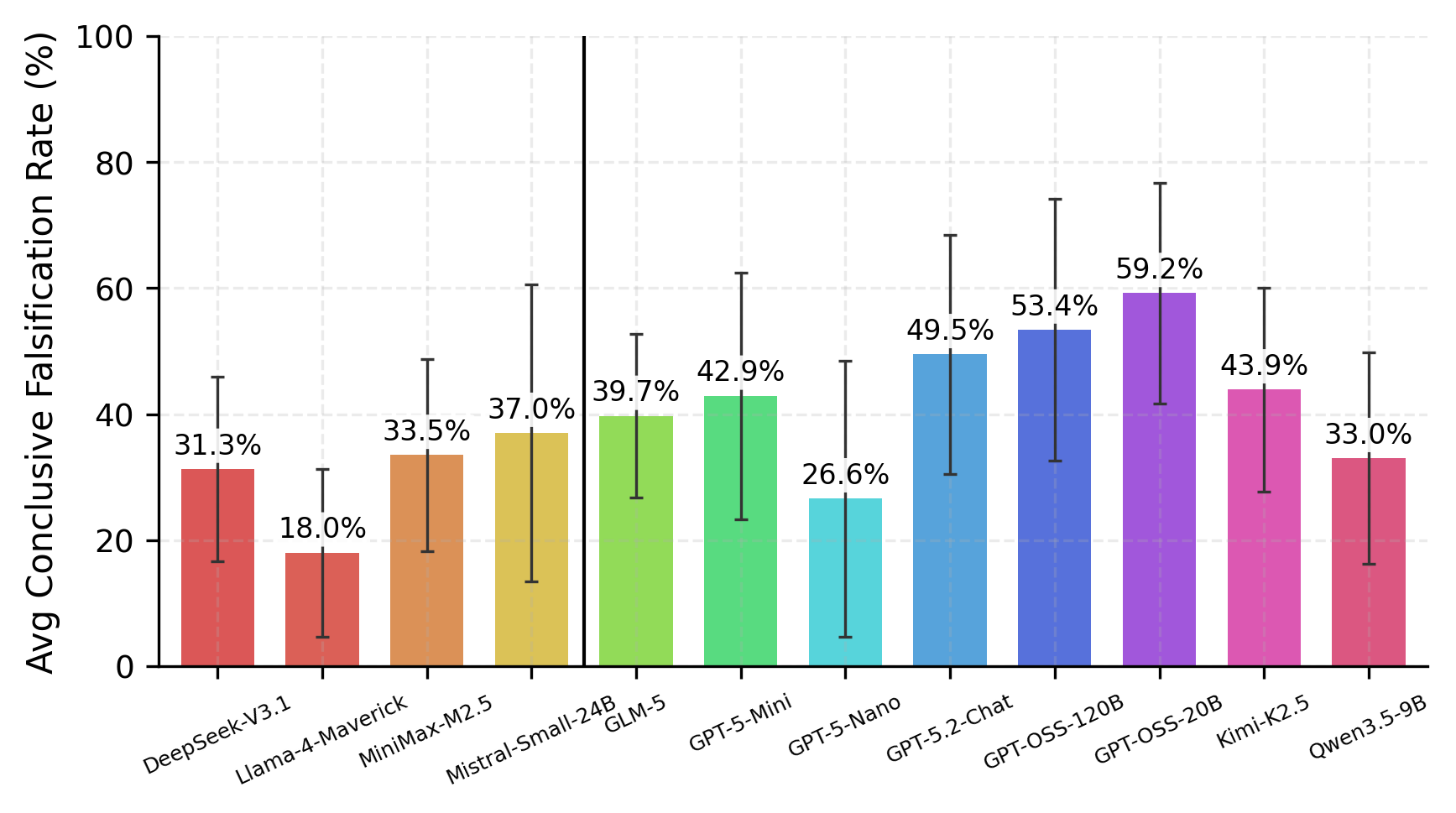}
    \\[6pt]
    \includegraphics[width=\linewidth]{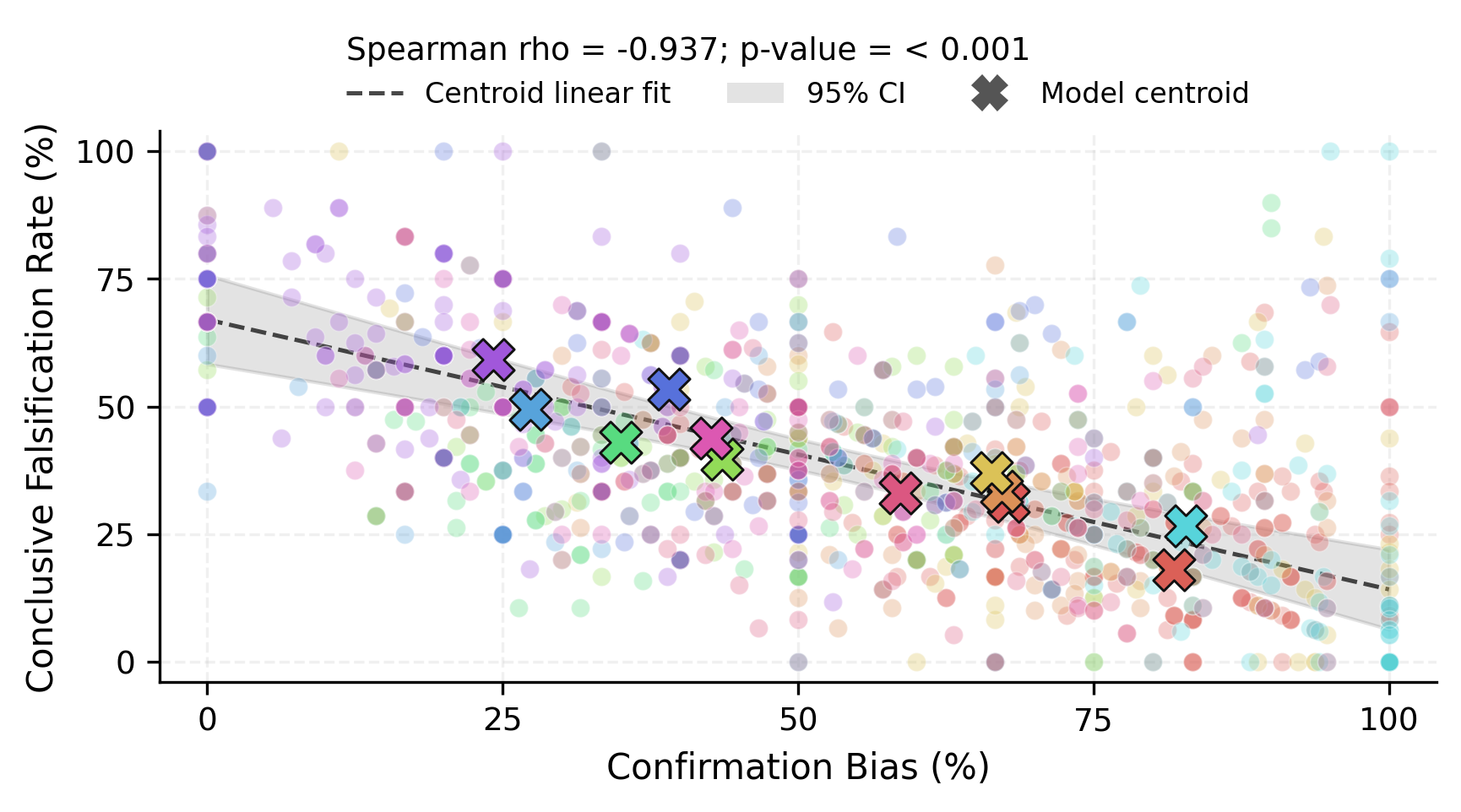}
    \caption{\textbf{Top:} Average conclusive falsification rate per model: the proportion of \texttt{Test} turns that yield a conclusive falsification given the relation between $H$ and $R$ at that turn. Models to the left of the vertical black line are \textbf{instruction-tuned} variants; models to the right are \textbf{reasoning} variants. \textbf{Bottom:} Scatter plot of Confirmation bias against conclusive falsification rate across all 12 models. The strong negative correlation (Spearman $\rho = -0.937$, $p < 0.001$) indicates that models that test more positively tend to reach fewer conclusive falsifications.}
    \label{fig:conclusive_and_bias}
\end{figure}

\paragraph{Confirmation bias vs.\ Conclusive Falsification Rate.}
Having access to the full annotation of the relation between $H$ and $R$ in each \texttt{Test} turn, we could check for each turn whether a positive or negative test would yield a falsification of $H$. Thus, we measured the average proportion of tests yielding a falsification of $H$ across games, called \emph{conclusive falsification rate} (Figure~\ref{fig:conclusive_and_bias}, top).
Reasoning models again outperformed instruction-tuned models on this measure, with GPT-OSS-20B (59.2\%), GPT-OSS-120B (53.4\%), and GPT-5.2-Chat (49.5\%) achieving the highest rates. The lowest rates were again produced by Llama-4-Maverick (18.0\%) and GPT-5-Nano (26.6\%). As explained earlier, negative testing is not always guaranteed to be the better strategy for falsifying a wrong hypothesis; however, we show here that it was indeed mostly the case in \textsc{FalsifyBench}: Figure~\ref{fig:conclusive_and_bias} (bottom) shows the Confirmation bias against the conclusive falsification rate across all models, revealing a strong negative correlation (Spearman $\rho = -0.937$, $p < 0.001$). The strength of this correlation supports the idea that testing negatively (i.e., having lower Confirmation bias) was a reliable proxy for reaching a conclusive falsification of wrong hypothesis in our framework.

\begin{figure*}[t!]
    \centering
    \begin{subfigure}[t]{0.48\linewidth}
        \centering
        \includegraphics[width=\linewidth]{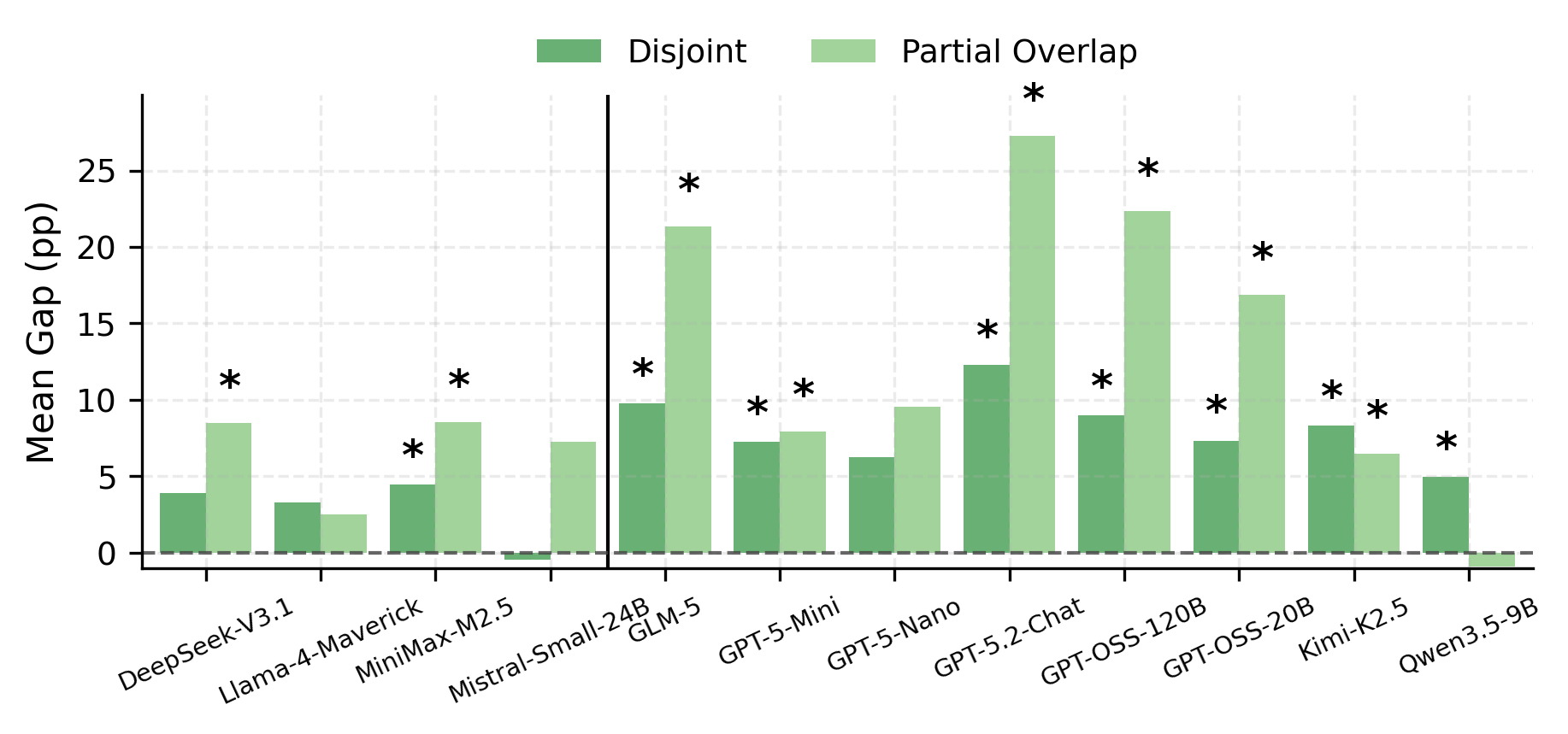}
    \end{subfigure}
    \hfill
    \begin{subfigure}[t]{0.48\linewidth}
        \centering
        \includegraphics[width=\linewidth]{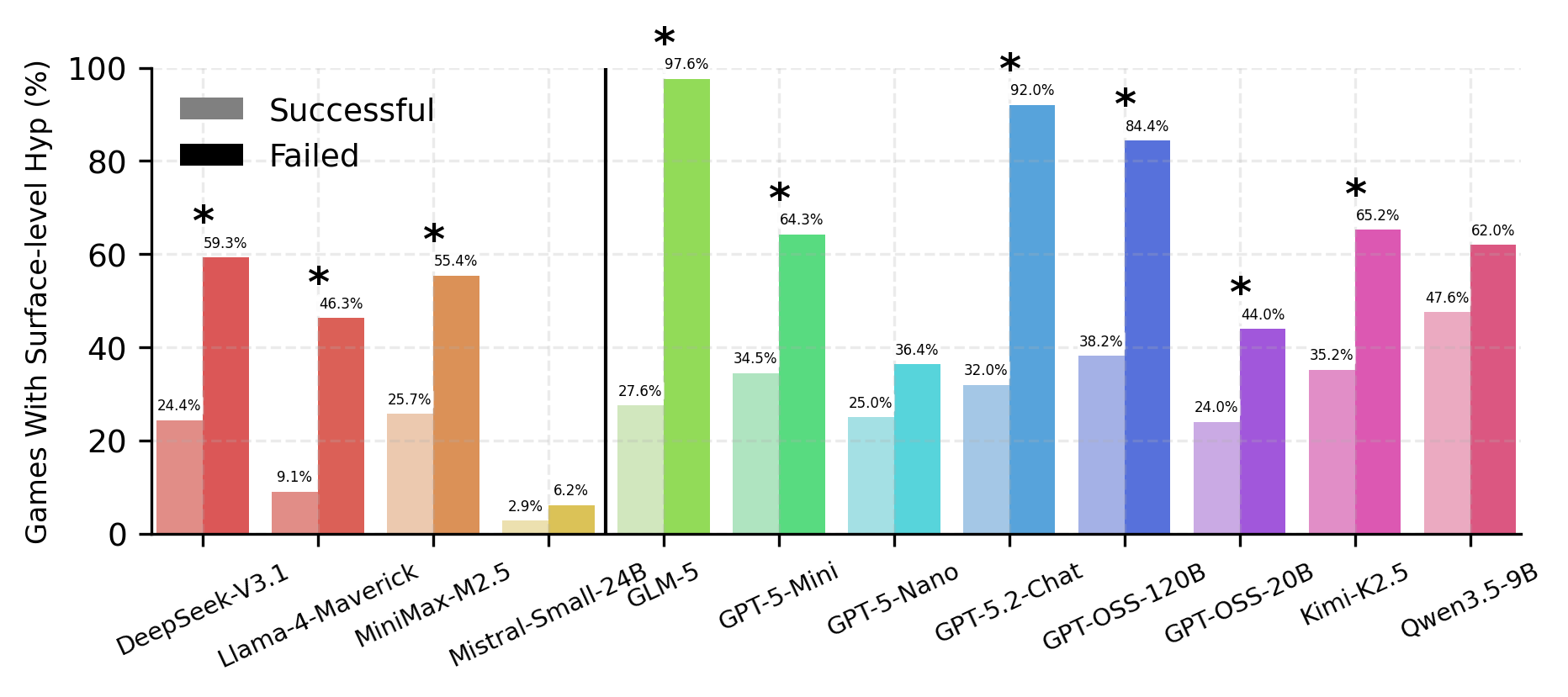}
    \end{subfigure}
    \caption{\textbf{Left:} Mean gap (in percentage points) in the proportion of Disjoint and Partial Overlap \texttt{Test} turns between failed and successful games, per model. Positive values indicate a configuration is more prevalent in failed games. Asterisks denote statistically significant differences in the median number of turns of each type between the two groups (Mann-Whitney U test, $p < 0.05$). \textbf{Right:} Percentage of successful and failed games in which at least one hypothesis generated across all turns was classified as focusing on surface-level linguistic features. Asterisks indicate models for which the proportion was significantly higher in failed than in successful games according to Fisher’s exact test. In both plots, models to the left of the vertical black line are \textbf{instruction-tuned} variants; models to the right are \textbf{reasoning} variants.}
    \label{fig:error_analysis}
\end{figure*}

\section{Failure Beyond Confirmation Bias}
\label{sec:error}

As shown in the previous sections, Confirmation bias is associated with failure in \textsc{FalsifyBench}. Beyond this bias, which is related to the testing strategy players adopted, failure may also be attributable to the type of hypotheses they generated. 

Failed games showed a systematically greater proportion of turns where the hypothesis was either disjoint from or partially overlapping with the target relative to successful games, indicating that models that failed did not follow a clear taxonomic path toward $R$.\footnote{We show a full per-model decomposition of relation distributions across successful and failed games in Appendix~\ref{app:error}.} Figure~\ref{fig:error_analysis} (left) quantifies this difference by reporting, for each model, the mean gap in the proportion of partially overlapping and disjoint turns between failed and successful games. Positive values indicate that a given configuration is more prevalent in failed games. Asterisks denote statistically significant differences in the median number of turns of each type (Mann-Whitney U test, $p < 0.05$). Nearly all models exhibited a significantly higher proportion of partially overlapping turns in failed games, with reasoning models showing particularly pronounced differences (GPT-5.2-Chat: $+27.0$ pp, GPT-OSS-120B: $+22.3$ pp, GLM-5: $+21.4$ pp). This suggests that in failed games models frequently formed hypotheses that were similar to $R$ but failed to converge on the correct rule. Disjoint turns were also significantly more frequent in failed games for most models, though the effect was smaller in magnitude, indicating that while fully off-track hypotheses did occur, they were less prevalent than partially overlapping ones.

While these results provide evidence that, in failed games, models often did not fully grasp the taxonomic structure of the hypothesis space, we also aimed to qualitatively characterize the kinds of hypotheses they generated when diverging from the correct path. A notable case involved a focus on surface-level features of the words used in the game (e.g.\ ``all items start with the letter a'' or ``all items have exactly three vowels in their names'') rather than on semantic properties (such as, ``is a fish'' or ``is a mammal''). To detect such cases, we built a heuristic classifier based on regular expressions.\footnote{Additional details regarding the regular expression patterns used to build the classifier are reported in Appendix~\ref{app:ling-feat}.} Figure~\ref{fig:error_analysis} (right) shows the percentage of games, both successful and failed, in which at least one hypothesis, across all game turns, was classified as focusing on surface-level features of items. Models for which a Fisher's exact test revealed a significantly higher proportion in failed games ($p < 0.05$) are denoted by an asterisk. This indicates that the tendency to form surface-level linguistic hypotheses was systematically associated with failure. Interestingly, for models with higher overall success rates, such as GLM-5, GPT-5.2-Chat and GPT-OSS-120B, nearly all failed games were characterized by the presence of at least one linguistic hypothesis (97.6\%, 92.0\% and 84.4\% respectively). In contrast, weaker models such as GPT-5-Nano and Qwen3.5-9B showed no significant difference between successful and failed games (36.4\% vs.\ 25.0\% and 62.0\% vs.\ 47.6\% respectively), suggesting that these models generally struggled with the task and did not exhibit this specific failure mode.

\section{Conclusion}

We introduced \textsc{FalsifyBench}, a benchmark for evaluating hypothesis-driven reasoning in LLMs through a semantic generalization of the Wason 2-4-6 task. Our evaluation of 12 models showed that reasoning models are generally stronger scientific thinkers than instruction-tuned models. However, no model came close to optimal performance, highlighting substantial room for improvement in the inductive reasoning capabilities of current LLMs. The primary driver of success was found to be the capacity for negative testing: models that actively seek to falsify their hypotheses consistently outperform those that seek to confirm them, and our turn-level analysis showed that disconfirmatory behavior is a reliable proxy for progress toward the correct hypothesis. 

Finally, we showed that failure is also tied to how models explored the hypothesis space. In failed games, models tended to drift away from the target, forming hypotheses that were either disjoint from or only partially overlapping with the target rule, and were significantly more likely to propose rules based on surface-level properties of words rather than semantic categories. 
%Together, these results establish \textsc{FalsifyBench} as a principled tool for studying and measuring hypothesis-driven reasoning in LLMs.

\section{Limitations}

Our analysis uses WordNet as a proxy for semantic taxonomic knowledge. Although WordNet is a well-established lexical resource created by expert lexicographers, it has several limitations. For example, its hierarchy reflects specific lexicographic decisions, it is English-centric, and it may contain basic semantic inconsistencies. We sought to overcome these limitations by focusing on concrete and fairly unambiguous superordinate categories.

Additionally, we collected human annotations for only one \texttt{Test} turn and one \texttt{Guess} turn per game. While this sampling strategy kept annotation effort manageable across 1,200 games, it forced us to operationalize oracle error as a binary variable. We acknowledge that this is a coarse approximation. Nevertheless, the high raw accuracy observed across models suggests that the probability of a randomly selected turn resulting in an incorrect oracle judgment is low. This means that our binary estimate is likely to have overestimated rather than underestimated the true oracle error rate, making our regression result conservative. Despite this, the effect of oracle error on player success remains negligible.

Finally, our analysis was inspired by the history of scientific discoveries, in which previously accepted hypotheses have been replaced by more comprehensive ones. We acknowledge, however, that this is only one aspect of the scientific process, and leave further exploration of this topic to future research. 

\bibliography{custom}

@article{wason1960failure,
  title={On the Failure to Eliminate Hypotheses in a Conceptual Task},
  author={Peter Cathcart Wason},
  journal={Quarterly Journal of Experimental Psychology},
  year={1960},
  volume={12},
  pages={129 - 140},
  url={https://api.semanticscholar.org/CorpusID:19237642}
}

@inproceedings{klayman1987confirmation,
  title={Confirmation, Disconfirmation, and Informa-tion in Hypothesis Testing},
  author={Joshua Klayman and Young Whee Ha},
  year={1987},
  url={https://api.semanticscholar.org/CorpusID:10853196}
}

@misc{banatt2024wilt,
      title={WILT: A Multi-Turn, Memorization-Robust Inductive Logic Benchmark for LLMs}, 
      author={Eryk Banatt and Jonathan Cheng and Skanda Vaidyanath and Tiffany Hwu},
      year={2024},
      eprint={2410.10998},
      archivePrefix={arXiv},
      primaryClass={cs.AI},
      url={https://arxiv.org/abs/2410.10998}, 
}

@misc{jhaveri2026falsify,
      title={Failing to Falsify: Evaluating and Mitigating Confirmation Bias in Language Models}, 
      author={Ayush Rajesh Jhaveri and Anthony GX-Chen and Ilia Sucholutsky and Eunsol Choi},
      year={2026},
      eprint={2604.02485},
      archivePrefix={arXiv},
      primaryClass={cs.CL},
      url={https://arxiv.org/abs/2604.02485}, 
}

@article{miller1995wordnet,
    author = {Miller, George A.},
    title = {WordNet: a lexical database for English},
    year = {1995},
    issue_date = {Nov. 1995},
    publisher = {Association for Computing Machinery},
    address = {New York, NY, USA},
    volume = {38},
    number = {11},
    issn = {0001-0782},
    url = {https://doi.org/10.1145/219717.219748},
    doi = {10.1145/219717.219748},
    journal = {Commun. ACM},
    month = nov,
    pages = {39–41},
    numpages = {3}
}

@inproceedings{
    jimenez2024swebench,
    title={{SWE}-bench: Can Language Models Resolve Real-world Github Issues?},
    author={Carlos E Jimenez and John Yang and Alexander Wettig and Shunyu Yao and Kexin Pei and Ofir Press and Karthik R Narasimhan},
    booktitle={The Twelfth International Conference on Learning Representations},
    year={2024},
    url={https://openreview.net/forum?id=VTF8yNQM66}
}

@inproceedings{
    jain2025livecodebench,
    title={LiveCodeBench: Holistic and Contamination Free Evaluation of Large Language Models for Code},
    author={Naman Jain and King Han and Alex Gu and Wen-Ding Li and Fanjia Yan and Tianjun Zhang and Sida Wang and Armando Solar-Lezama and Koushik Sen and Ion Stoica},
    booktitle={The Thirteenth International Conference on Learning Representations},
    year={2025},
    url={https://openreview.net/forum?id=chfJJYC3iL}
}

@inproceedings{
    wang2024mmlupro,
    title={{MMLU}-Pro: A More Robust and Challenging Multi-Task Language Understanding Benchmark},
    author={Yubo Wang and Xueguang Ma and Ge Zhang and Yuansheng Ni and Abhranil Chandra and Shiguang Guo and Weiming Ren and Aaran Arulraj and Xuan He and Ziyan Jiang and Tianle Li and Max Ku and Kai Wang and Alex Zhuang and Rongqi Fan and Xiang Yue and Wenhu Chen},
    booktitle={The Thirty-eight Conference on Neural Information Processing Systems Datasets and Benchmarks Track},
    year={2024},
    url={https://openreview.net/forum?id=y10DM6R2r3}
}

@inproceedings{rein2024gpqa,
      title={{GPQA}: A Graduate-Level Google-Proof Q\&A Benchmark},
      author={David Rein and Betty Li Hou and Asa Cooper Stickland and Jackson Petty and Richard Yuanzhe Pang and Julien Dirani and Julian Michael and Samuel R. Bowman},
      booktitle={First Conference on Language Modeling},
      year={2024},
      url={https://openreview.net/forum?id=Ti67584b98}
}

@article{phan2025lastexam,
      title = {A benchmark of expert-level academic questions to assess {AI} capabilities},
      author = {{Center for AI Safety} and {Scale AI} and {HLE Contributors Consortium}},
      journal = {Nature},
      volume = {649},
      pages = {1139--1146},
      year = {2026},
      doi = {10.1038/s41586-025-09962-4},
      eprint = {2501.14249},
      archivePrefix = {arXiv},
      primaryClass = {cs.LG},
      url = {https://arxiv.org/abs/2501.14249}
}

@inproceedings{
    mialon2024gaia,
    title={{GAIA}: a benchmark for General {AI} Assistants},
    author={Gr{\'e}goire Mialon and Cl{\'e}mentine Fourrier and Thomas Wolf and Yann LeCun and Thomas Scialom},
    booktitle={The Twelfth International Conference on Learning Representations},
    year={2024},
    url={https://openreview.net/forum?id=fibxvahvs3}
}

@misc{
    wei2025browsecomp,
    title={BrowseComp: A Simple Yet Challenging Benchmark for Browsing Agents}, 
    author={Jason Wei and Zhiqing Sun and Spencer Papay and Scott McKinney and Jeffrey Han and Isa Fulford and Hyung Won Chung and Alex Tachard Passos and William Fedus and Amelia Glaese},
    year={2025},
    eprint={2504.12516},
    archivePrefix={arXiv},
    primaryClass={cs.CL},
    url={https://arxiv.org/abs/2504.12516}, 
}

@misc{
    glazer2025frontiermath,
    title={FrontierMath: A Benchmark for Evaluating Advanced Mathematical Reasoning in AI}, 
    author={Elliot Glazer and Ege Erdil and Tamay Besiroglu and Diego Chicharro and Evan Chen and Alex Gunning and Caroline Falkman Olsson and Jean-Stanislas Denain and Anson Ho and Emily de Oliveira Santos and Olli Järviniemi and Matthew Barnett and Robert Sandler and Matej Vrzala and Jaime Sevilla and Qiuyu Ren and Elizabeth Pratt and Lionel Levine and Grant Barkley and Natalie Stewart and Bogdan Grechuk and Tetiana Grechuk and Shreepranav Varma Enugandla and Mark Wildon},
    year={2025},
    eprint={2411.04872},
    archivePrefix={arXiv},
    primaryClass={cs.AI},
    url={https://arxiv.org/abs/2411.04872}, 
}

@inproceedings{
    mirzadeh2025gsmsymbolic,
    title={{GSM}-Symbolic: Understanding the Limitations of Mathematical Reasoning in Large Language Models},
    author={Seyed Iman Mirzadeh and Keivan Alizadeh and Hooman Shahrokhi and Oncel Tuzel and Samy Bengio and Mehrdad Farajtabar},
    booktitle={The Thirteenth International Conference on Learning Representations},
    year={2025},
    url={https://openreview.net/forum?id=AjXkRZIvjB}
}

@misc{
    gottweis2025aicoscientist,
    title={Towards an AI co-scientist}, 
    author={Juraj Gottweis and Wei-Hung Weng and Alexander Daryin and Tao Tu and Anil Palepu and Petar Sirkovic and Artiom Myaskovsky and Felix Weissenberger and Keran Rong and Ryutaro Tanno and Khaled Saab and Dan Popovici and Jacob Blum and Fan Zhang and Katherine Chou and Avinatan Hassidim and Burak Gokturk and Amin Vahdat and Pushmeet Kohli and Yossi Matias and Andrew Carroll and Kavita Kulkarni and Nenad Tomasev and Yuan Guan and Vikram Dhillon and Eeshit Dhaval Vaishnav and Byron Lee and Tiago R D Costa and José R Penadés and Gary Peltz and Yunhan Xu and Annalisa Pawlosky and Alan Karthikesalingam and Vivek Natarajan},
    year={2025},
    eprint={2502.18864},
    archivePrefix={arXiv},
    primaryClass={cs.AI},
    url={https://arxiv.org/abs/2502.18864}, 
}

@misc{
    lu2024aiscientistfullyautomated,
    title={The AI Scientist: Towards Fully Automated Open-Ended Scientific Discovery}, 
    author={Chris Lu and Cong Lu and Robert Tjarko Lange and Jakob Foerster and Jeff Clune and David Ha},
    year={2024},
    eprint={2408.06292},
    archivePrefix={arXiv},
    primaryClass={cs.AI},
    url={https://arxiv.org/abs/2408.06292}, 
}

@inproceedings{
    li2025mirage,
    title={{MIRAGE}: Evaluating and Explaining Inductive Reasoning Process in Language Models},
    author={Jiachun Li and Pengfei Cao and Zhuoran Jin and Yubo Chen and Kang Liu and Jun Zhao},
    booktitle={The Thirteenth International Conference on Learning Representations},
    year={2025},
    url={https://openreview.net/forum?id=tZCqSVncRf}
}

@inproceedings{
    jin2026evaluating,
    title={Evaluating the Inductive Abilities of Large Language Models: Why Chain-of-Thought Reasoning Sometimes Hurts More Than Helps},
    author={Haibo Jin and Peiyan Zhang and Man Luo and Haohan Wang},
    booktitle={The Thirty-ninth Annual Conference on Neural Information Processing Systems},
    year={2026},
    url={https://openreview.net/forum?id=yRxX01oRIi}
}

@misc{louapre2026gameofscience,
  title={Can LLMs Play the Game of Science?},
  author={David Louapre},
  year={2026},
  url={https://huggingface.co/spaces/huggingface/eleusis-benchmark}
}

@article{chollet2019arc,
  author       = {Fran{\c{c}}ois Chollet},
  title        = {On the Measure of Intelligence},
  journal      = {CoRR},
  volume       = {abs/1911.01547},
  year         = {2019},
  url          = {http://arxiv.org/abs/1911.01547},
  eprinttype   = {arXiv},
  eprint       = {1911.01547},
  bibsource    = {dblp computer science bibliography, https://dblp.org}
}

@inproceedings{bird2004nltk,
    title = "{NLTK}: The Natural Language Toolkit",
    author = "Bird, Steven  and
      Loper, Edward",
    booktitle = "Proceedings of the {ACL} Interactive Poster and Demonstration Sessions",
    month = jul,
    year = "2004",
    address = "Barcelona, Spain",
    publisher = "Association for Computational Linguistics",
    url = "https://aclanthology.org/P04-3031/",
    pages = "214--217"
}

@InProceedings{lake2018scan,
    title = 	 {Generalization without Systematicity: On the Compositional Skills of Sequence-to-Sequence Recurrent Networks},
    author =       {Lake, Brenden and Baroni, Marco},
    booktitle = 	 {Proceedings of the 35th International Conference on Machine Learning},
    pages = 	 {2873--2882},
    year = 	 {2018},
    editor = 	 {Dy, Jennifer and Krause, Andreas},
    volume = 	 {80},
    series = 	 {Proceedings of Machine Learning Research},
    month = 	 {10--15 Jul},
    publisher =    {PMLR},
    url = 	 {https://proceedings.mlr.press/v80/lake18a.html}
}

@inproceedings{wei2022cot,
    author = {Wei, Jason and Wang, Xuezhi and Schuurmans, Dale and Bosma, Maarten and Ichter, Brian and Xia, Fei and Chi, Ed H. and Le, Quoc V. and Zhou, Denny},
    title = {Chain-of-thought prompting elicits reasoning in large language models},
    year = {2022},
    isbn = {9781713871088},
    publisher = {Curran Associates Inc.},
    address = {Red Hook, NY, USA},
    booktitle = {Proceedings of the 36th International Conference on Neural Information Processing Systems},
    articleno = {1800},
    numpages = {14},
    location = {New Orleans, LA, USA},
    series = {NIPS '22}
}
 
\appendix

\section{When Can a Hypothesis be Falsified?}
\label{app:falsification}

We provide a detailed analysis of when a test yields a falsification depending on the relation between the current hypothesis $H$ and the target rule $R$. Following \citet{klayman1987confirmation}, we classify any test $x$ along two binary dimensions: whether $x \in H$ (the test is \emph{positive} with respect to the hypothesis) and whether $x \in R$ 
(the test is \emph{true} with respect to the target). A test within $H$ is a \emph{positive} test; a test outside $H$ is a \emph{negative} test. Whether the oracle confirms or rejects it determines whether the test is \emph{true} or \emph{false}. This yields four cases, summarized in Table~\ref{tab:cells}, each with a distinct informational value for hypothesis revision.

\begin{table}[h]
\centering
\small
\begin{tabular}{lcc}
\toprule
 & \textbf{$x \in R$} & \textbf{$x \notin R$} \\
\midrule
\textbf{$x \in H$}    & True positive          
                       & False positive \\
\textbf{$x \notin H$} & False negative        
                       & True negative \\
\bottomrule
\end{tabular}
\caption{Classification of a test $x$ by membership in $H$ and $R$.}
\label{tab:cells}
\end{table}

In the following sections, we discuss in each possible relation between $H$ and $R$ which of the four quadrants yield a conclusive falsification and which do not.

\subsection*{Case 1: $H = R$ (Identical)}

When $H$ is identical to $R$, both false positives and false negatives are impossible. Every member of $H$ is in $R$ (so no test within $H$ can receive ``Do not conform''), and no member outside $H$ is in $R$ (so no test outside $H$ can 
receive ``Conform''). Trivially, holding the correct hypothesis means that it cannot be falsified and only a \texttt{Guess} action would end the game.

\begin{figure}[h]
    \centering
    \includegraphics[width=0.5\linewidth]{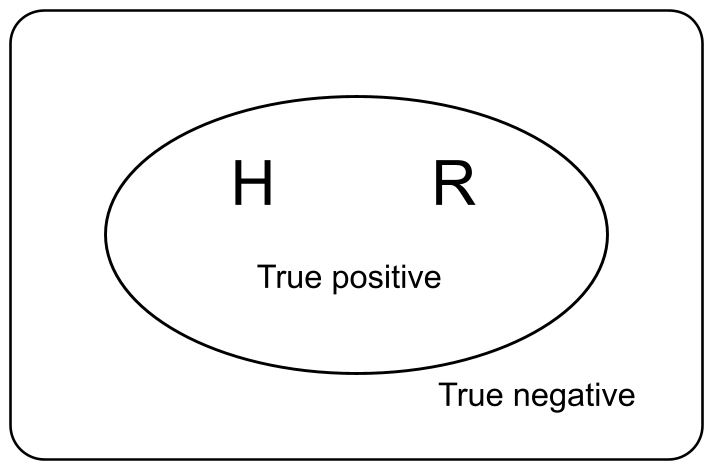}
\end{figure}

\subsection*{Case 2: $H \subset R$ (H too narrow)}
When $H$ is a strict subset of $R$, every member of $H$ is automatically in $R$, making false positives impossible: tests within $H$ always receive ``Conform''. The only route to falsification is a false negative -- testing outside $H$ and receiving ``Conform'', revealing that $R$ extends beyond $H$ and forcing upward revision. Crucially, not every negative test achieves this: testing outside $H$ and receiving ``Do not conform'' (true negative) is equally consistent with $H$ and provides no new information. A player relying exclusively on positive tests is therefore confined to true positives and can never falsify $H$. This is the dominant relation in \textsc{FalsifyBench}.

\begin{figure}[h]
    \centering
    \includegraphics[width=0.5\linewidth]{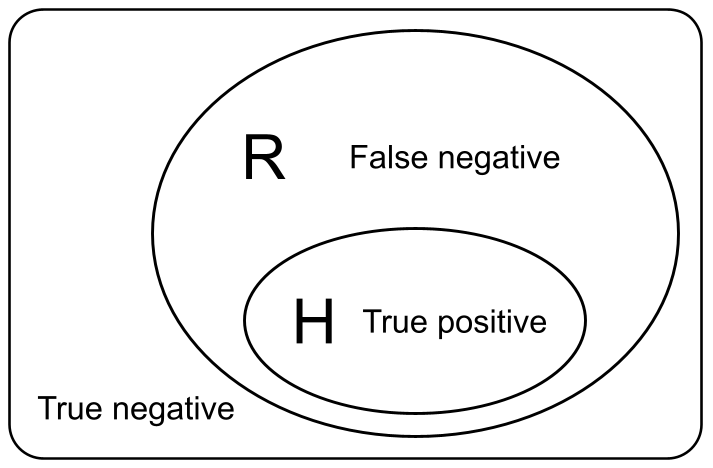}
\end{figure}

\subsection*{Case 3: $H \supset R$ (H too broad)}
When $H$ is strictly broader than $R$, every member of $R$ is inside $H$, making false negatives impossible: no test outside $H$ can receive ``Conform''. The only route to conclusive falsification is a false positive -- testing within $H$ and receiving ``Do not conform'', revealing that $H$ must be narrowed. Notably, a positive test strategy is the correct approach in this configuration: testing within $H$ is the only way to a falsification. Crucially, not every positive test achieves this: testing within $H$ and receiving ``Conform'' (true positive) is equally consistent with $H$ and provides no reason to discard it.

\begin{figure}[h]
    \centering
    \includegraphics[width=0.5\linewidth]{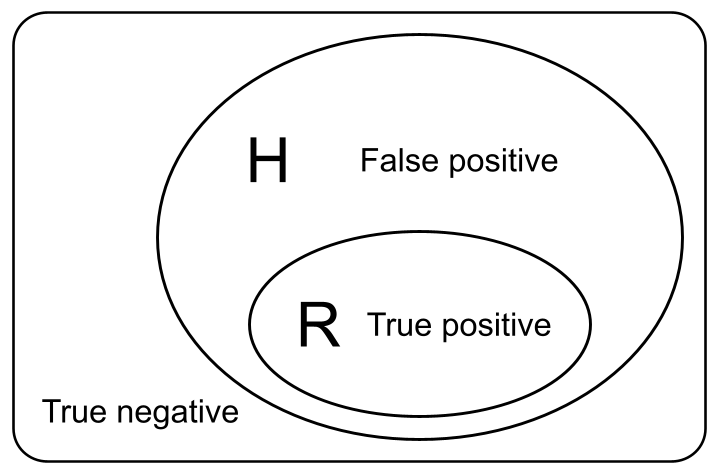}
\end{figure}

\subsection*{Case 4: $H$ and $R$ partially overlap}
When $H$ and $R$ partially overlap but neither contains the other, both false positives and false negatives are accessible. A test within $H$ but outside $R$ yields ``Do not conform'' (false positive, falsification): only a subset of $H$ is compatible with $R$, revealing that $H$ must be revised to exclude that region. A test within $R$ but outside $H$ yields ``Conform'' (false negative, falsification): $R$ extends beyond $H$, revealing that $H$ must be revised to include that region. Thus, two ways of falsification are available in this configuration. However, two non-conclusive outcomes are also possible: testing within both $H$ and $R$ yields ``Conform'' (true positive, non-conclusive), and testing outside both $H$ and $R$ yields ``Do not conform'' (true negative, non-conclusive). In both cases the feedback is consistent with $H$ and provides no pressure to revise it in either direction.

\begin{figure}[h]
    \centering
    \includegraphics[width=0.5\linewidth]{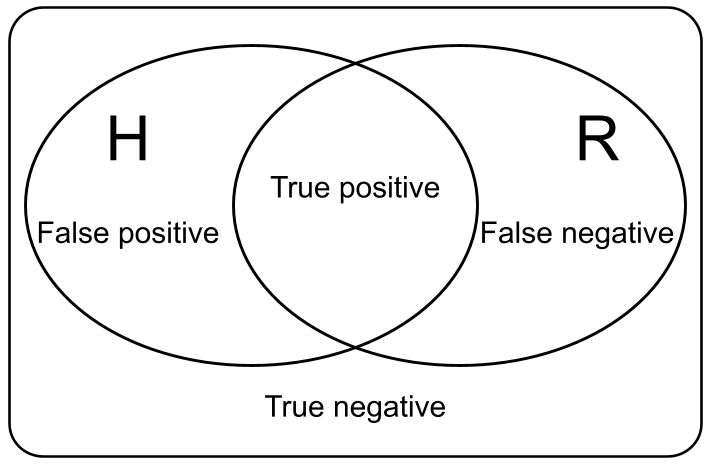}
\end{figure}

\subsection*{Case 5: $H$ and $R$ disjoint}
When $H$ and $R$ share no members, every test within $H$ yields ``Do not conform'' (false positive, falsification): $H$ is entirely incompatible with $R$ and must be discarded. Every test within $R$ but outside $H$ yields ``Conform'' (false negative, falsification): $R$ extends into a region entirely outside $H$, and $H$ needs to be revised. The only non-conclusive outcome is testing outside both $H$ and $R$, which yields ``Do not conform'' (true negative, non-conclusive): the feedback is consistent with $H$ and provides no information about how to revise it. Finally, true positives are impossible in this configuration, since $H$ and $R$ share no members.

\begin{figure}[h]
    \centering
    \includegraphics[width=0.5\linewidth]{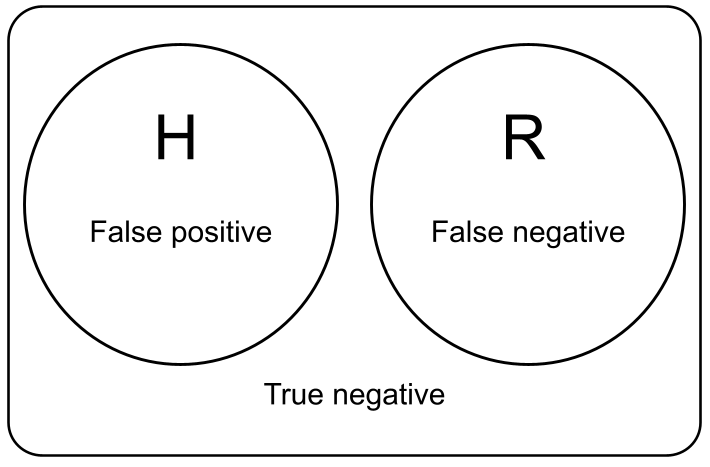}
\end{figure}

\section{Game Construction Details}
\label{app:games}

We build game instances using the WordNet taxonomy, where each node is a semantic category and its descendants represent increasingly specific subcategories. As we explained in Section~\ref{sec:exp_setup}, we started with seven high-level WordNet categories representing concrete concepts as candidate target rules: ``animal'', ``artifact'', ``body part'', ``food'', ``location'', ``plant'', and ``worker''. However, many of the possible descendants of these categories involve low-frequency vocabulary. Hence, we manually filtered the subcategory pool, selecting only five target categories: ``animal'', ``artifact'', ``body part'', ``food'' and ``plant''.

Since subcategories of these target categories can come from different depths in the taxonomy (e.g. ``feline'' is deeper than ``mammal''), we first divided the subcategories into two groups based on their taxonomic distance from the targets. The \emph{close} group has a distance of at most four edges, and the \emph{deep} group exceeds four edges. To ensure diversity in the sampling categories for each game, we sampled triples from the close and deep categories in equal proportions. Table~\ref{tab:properties} lists all the selected sampling categories for the close and deep categories, respectively, grouped by target rule.

\begin{table*}[ht]
\centering
\small
\begin{tabularx}{\textwidth}{lXX}
\toprule
\textbf{Target rule} & \textbf{Close categories ($\leq4$)} & \textbf{Deep categories ($>4$)} \\
\midrule

animal &
aquatic vertebrate, bird, mammal, vertebrate &
Australian magpie, New World blackbird, New World flycatcher,
New World oriole, New World sparrow, New World vulture,
New World warbler, Old World vulture, Old World warbler,
bear, big cat, bird of prey, bony fish, cat, dog, eagle,
falcon, feline, fox, goose, hawk, lion, owl, shark,
sparrow, sunfish, swallow, tiger, vulture, whale, wolf \\

\addlinespace

artifact &
bag, clothing, container, electrical device, electronic device,
musical instrument, utensil &
aircraft, airplane, bioweapon, blanket, chamber, church,
cooling system, cutter, cutting implement, engine, guitar,
hammer, horn, lamp, measuring instrument, missile, motor,
painting, pedal, pottery, propeller, robe, ship,
stringed instrument, tube, valve, vessel, warship,
weapon, wind instrument \\

\addlinespace

body part &
internal organ, muscle, reproductive organ, sense organ &
artery, blood vessel, bone, conjunctiva, facial muscle,
mucous membrane, nose, ocular muscle, skeletal muscle,
vein, venous sinus, vesicle \\

\addlinespace

food &
appetizer, bread, cake, dessert, meal, pastry, pie, soup &
bacon, beefsteak, biscuit, broth, candy, chicken, cocktail,
condiment, conserve, cookout, cornbread, dip, fillet,
flatbread, frozen dessert, fudge, goulash, ice cream,
muffin, pudding, sugar, sweet roll, tart, toast \\

\addlinespace

plant &
angiosperm, aquatic plant, gymnosperm, herb, tree, woody plant &
anemone, anise tree, arborvitae, barberry, cedar, clematis,
columbine, conifer, custard apple, cycad, cypress, fir,
fruit tree, hemlock, larch, laurel, magnolia, pine,
silver fir, spruce, white pine, wildflower \\

\bottomrule
\end{tabularx}

\caption{Selected sampling categories grouped by target rule and taxonomic
distance from the target. Close categories have distance $\leq4$ in the
WordNet taxonomy, while deep categories have distance $>4$.}
\label{tab:properties}
\end{table*}

\section{Game Engine Details}
\label{app:game_engine}

\subsection{Prompts}
 
The prompts we used verbatim in all experiments are shown in Tables~\ref{box:player}--\ref{box:annotation}. In each prompt, double curly-brace placeholders (e.g.\ \verb|{{max_turns}}|) are filled before each game. The player prompts are shown in Table~\ref{box:player}; the oracle prompts are shown in Table~\ref{box:oracle}. Additionally, Table~\ref{box:annotation} shows the prompt used for the turn level annotation of the relations between a current hypothesis and the target rule.
 
\subsection{Agent Roles and Action Space}
 
Each game began with exactly three positive examples presented to the player. The player was never shown the target rule $R$ or the sampling category $S$ directly; it observed only the three example items and the system prompt explaining the rules of the game (see Table~\ref{box:player}).

The interaction proceeded as a sequence of turns between the player and the oracle. The player was responsible for proposing actions aimed at discovering the hidden property, while the oracle evaluated these actions and provided feedback according to the player's last action.

On each turn, the player performed exactly one of two actions:

\begin{itemize}
\item \textbf{Test:} propose a triple of items together with a hypothesis string and a reasoning trace. The action must be well-formed JSON matching the schema specified in the system prompt; malformed or schema-invalid submissions trigger a retry (see Appendix \ref{app:retries}) and are not recorded as turns.
\item \textbf{Guess:} submit a single property string as a final answer. A correct guess returns ``Correct'' and ends the game; an incorrect guess returns ``Incorrect'', after which the player can continue with additional \texttt{Test} or \texttt{Guess} actions.
\end{itemize}

\noindent After receiving a valid player action, the oracle performed the following actions:

\begin{itemize}
\item \textbf{Feedback generation}: for a \texttt{Test} action, determine whether the proposed items satisfy the target rule $R$ and return the string ``Conform'' if they do and ``Do not conform'' otherwise; for a \texttt{Guess} action, determine whether the proposed property is semantically equivalent to $R$ by returning the string ``Correct'' if it does and ``Incorrect'' otherwise.
\item \textbf{Strategy classification}: after every \texttt{Test} action, classify the player's testing strategy as \emph{positive} if the player tests examples expected to satisfy its current hypothesis, and \emph{negative} if it tests examples outside the hypothesis in order to challenge it and explore broader alternatives. This step was performed only for later analysis and did not affect gameplay directly.
\end{itemize}

\noindent A game turn was defined as one accepted player action together with the oracle's response. Invalid submissions and retry attempts did not increment the turn counter.
 
\subsection{Retry Mechanism}
\label{app:retries}
 
The game engine employed two independent retry mechanisms to handle failures in producing valid JSON outputs, and none of these retry attempts were recorded as game turns. For player actions, if the engine received malformed JSON or an action that violated the schema, it re-prompted within the same turn requesting strict JSON output; after the first failure, up to three additional retries were permitted. For oracle judgments, each method was allowed up to two attempts: if the first response could not be parsed as JSON, the engine re-issued the prompt with an appended instruction (``IMPORTANT: Return ONLY a JSON object'') and then attempted to parse the second response. This numeber of retries was sufficient to complete all games with all tested models.
 
\subsection{Memory and statefulness.}
\label{app:memory}
During the game interactions, the player was \emph{stateful}: its conversation history accumulated all prior oracle feedback and its own previous actions. In contrast, the oracle was \emph{stateless}: each call to the oracle was performed using a newly initialized prompt that provided only the elements necessary to collect the current judgement, with no access to information from earlier turns.
 
\section{Experiment Details}
\label{app:exp_details}

\subsection{LLMs Details}

The following are the model identifiers of all evaluated LLMs: 

\begin{center}
{
\scriptsize
\label{tab:models}
\begin{tabular}{l}
\toprule
\textbf{Identifier} \\
\midrule
\texttt{gpt-5.2-chat} \\
\texttt{gpt-5-mini} \\
\texttt{gpt-5-nano} \\
\texttt{openai/gpt-oss-120b} \\
\texttt{openai/gpt-oss-20b} \\
\texttt{deepseek-ai/DeepSeek-V3.1} \\
\texttt{Qwen/Qwen3.5-9B} \\
\texttt{MiniMaxAI/MiniMax-M2.5} \\
\texttt{meta-llama/Llama-4-Maverick-17B-128E-Instruct-FP8} \\
\texttt{mistralai/Mistral-Small-24B-Instruct-2501} \\
\texttt{moonshotai/Kimi-K2.5} \\
\texttt{zai-org/GLM-5} \\
\bottomrule
\end{tabular}
}
\end{center}

\noindent Among these models, \texttt{gpt-5.2-chat} was used via the Azure OpenAI API, while \texttt{gpt-5-mini} and \texttt{gpt-5-nano} were used via the standard OpenAI API. All other models were evaluated using the TogetherAI API. 

When used in the oracle role, reasoning models were run with either low reasoning or no reasoning, depending on the model. When used in the player role, reasoning models were run at a medium level of effort. When the player failed to produce a valid JSON action, low reasoning effort was requested to speed up generation during retries. Regarding temperature and other sampling parameters, we used the provider defaults for all instruction-tuned models. For reasoning models, parameters were adapted according to provider-specific instructions depending on whether the model was used in reasoning or non-reasoning mode. The total cost of running all experiments across all models amounted to approximately \$150.

\begin{table*}[t!]
\centering
\small
\begin{tabular}{ll}
\toprule
\textbf{Rule} & \textbf{Trigger keywords / pattern} \\
\midrule
\textsc{syllable} & \textit{syllable}, \textit{syllables}, \textit{syllabic} \\
\textsc{letter} & \textit{letter}, \textit{letters} \\
\textsc{word\_count} & \textit{single-word}, \textit{two word}, \textit{multiple words}, \ldots \\
\textsc{prefix\_suffix} & \textit{prefix}, \textit{suffix}, \textit{prefixes}, \textit{suffixes} \\
\textsc{vowel\_consonant} & \textit{vowel(s)}, \textit{consonant(s)} \\
\textsc{spelling} & \textit{spelling}, \textit{spelled}, \textit{spelt}, \textit{orthography}, \textit{orthographic} \\
\textsc{pronunciation} & \textit{pronounce}, \textit{pronounced}, \textit{pronunciation}, \textit{phonetic}, \textit{rhyme}, \textit{rhymes}, \textit{rhyming} \\
\textsc{naming} & \textit{common name}, \textit{scientific name}, \textit{latin name}, \textit{names have} \\
\textsc{alphabetic} & \textit{alphabet}, \textit{alphabetic}, \textit{alphabetical} \\
\textsc{initial\_final\_unit} & \textit{first letter}, \textit{last word}, \textit{initial syllable}, \textit{final word}, \ldots \\
\textsc{meta\_ling.\ struct.} & \{\textit{starts}, \textit{begins}, \textit{ends}, ...\} $\times$ \{\textit{letter}, \textit{word}, \textit{name}, \textit{syllable}, ...\} \\
\bottomrule
\end{tabular}
\caption{Patterns used in the surface-level linguistic feature classifier. Each rule is triggered when the hypothesis text matches the listed keywords or pattern. A single hypothesis may match multiple rules simultaneously.}
\label{tab:ling-rules}
\end{table*}

\subsection{Mixed-Effects Logistic Regression Details}
\label{app:mixed-effects}

The experimental design of \textsc{FalsifyBench} assigned each LLM to both the player and oracle roles. As reported in Section~\ref{sec:results}, we performed a game-level analysis to estimate how player strategy and oracle feedback relate to player success.
Since we had 12 LLMs and each played 100 games, we obtained a total of 1,200 game-level observations. We fit a mixed-effects logistic regression using variational Bayes inference of the form:

\begin{equation}
\begin{aligned}
\text{logit}\,P(\text{success}_{ij}) =\;& \alpha \\
&+ \beta_1 \cdot \text{confirmation\_bias}_{ij} \\
&+ \beta_2 \cdot \text{oracle\_error}_{ij} \\
&+ u_j
\end{aligned}
\end{equation}

\noindent where confirmation bias is a continuous variable in the interval $[0,1]$, oracle error is a binary variable in $\{0,1\}$, $i$ indexes games, $j$ indexes models, and $u_j \sim \mathcal{N}(0,\sigma_u^2)$ is an LLM-level random intercept. The model was fit using variational Bayes inference with \texttt{BinomialBayesMixedGLM} from the \texttt{statsmodels} Python package. We report posterior means, posterior standard deviations, and implied odds ratios for all fixed effects as follows:

\begin{tcolorbox}[colback=white, colframe=white, boxrule=0pt, sharp corners, left=2pt, right=2pt, top=2pt, bottom=2pt]
\textbf{Fixed Effects}\\
\texttt{Intercept: 1.955 (SD 0.066), OR 7.067}\\
\texttt{Confirmation bias: -4.148 (SD 0.116), OR 0.016}\\
\texttt{Oracle error: -0.123 (SD 0.147), OR 0.885}\\
\textbf{Random Effects}\\
\texttt{LLM SD ($\hat{\sigma}_u$): 0.553}
\end{tcolorbox}

\noindent In this analysis the posterior standard deviation is used as a measure of estimation uncertainty. An approximate 95\% credible interval is obtained using a normal approximation $\mu \pm 1.96\sigma$.

Confirmation bias was a strong negative predictor of success. Since Confirmation bias ranges from 0 (no Confirmation bias) to 1 (maximum Confirmation bias), the corresponding odds ratio of $0.016$ reflects the effect of moving across the full range of the variable, reducing the odds of success to approximately $1/63$ of their original value. The 95\% credible interval for the log-odds coefficient is $[-4.148 \pm 1.96 \times 0.116] = [-4.375, -3.921]$, entirely below zero, confirming a robust negative effect.

By contrast, oracle error had a negligible and highly uncertain effect. The implied odds ratio of $0.885$ corresponds to an approximate 11.5\% reduction in the odds of success when an oracle error is present, though this effect is small and the credible interval $[-0.123 \pm 1.96 \times 0.147] = [-0.411, 0.165]$ spans zero, indicating that the effect is not credibly different from zero.

The random-effect standard deviation of $\hat{\sigma}_u = 0.553$ indicates moderate residual between-model variability not explained by the two predictors. This is expected, as model identity captures additional factors that generally determine underlying LLM quality, such as model size, amount of training data, and post-training procedures, none of which are explicitly modeled in the fixed effects.

\subsection{Linguistic Feature Classifier Details}
\label{app:ling-feat}

\begin{figure}[t]
    \centering
    \includegraphics[width=0.97\linewidth]{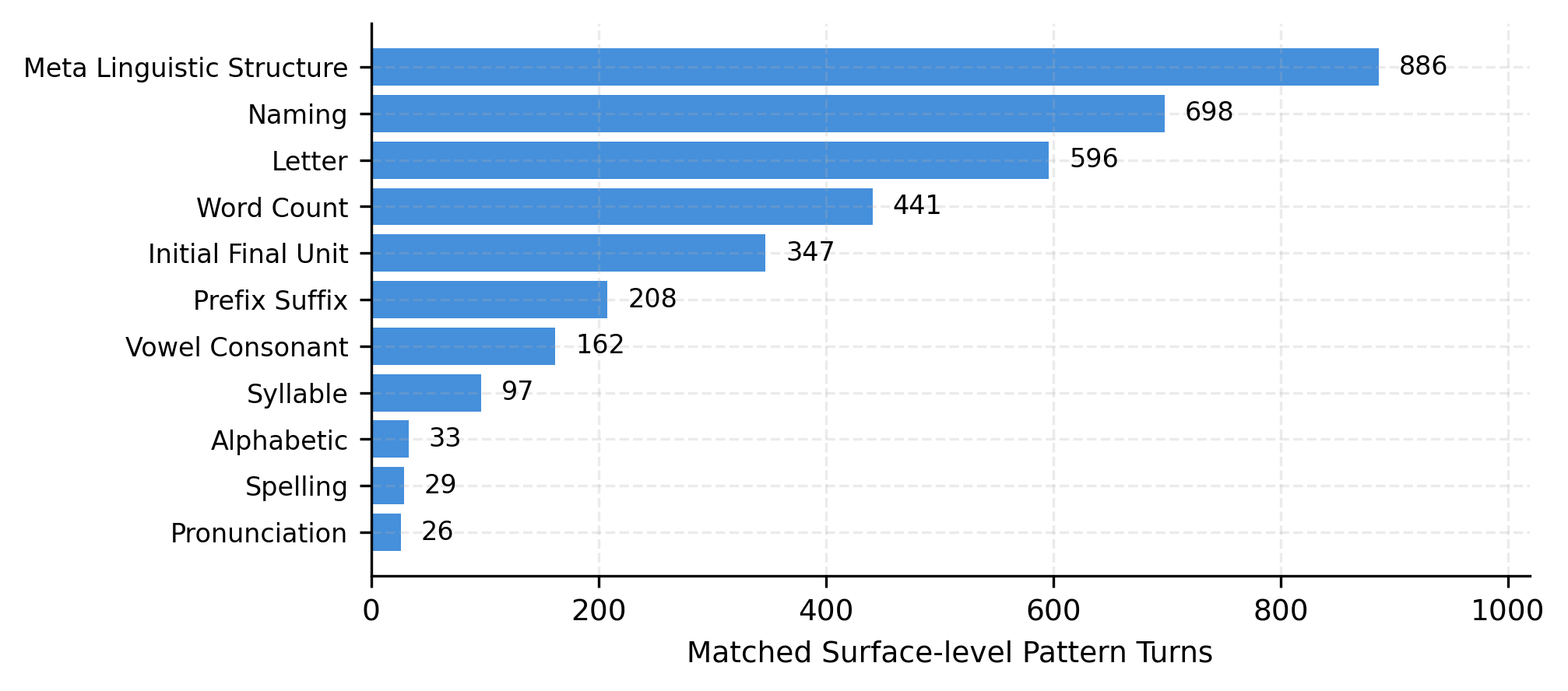}
    \caption{Raw match count between surface-level linguistic patterns on game turns.}
    \label{fig:rule_count}
\end{figure}

To detect hypotheses that focus on surface-level linguistic properties of words rather than their semantic content, we designed a heuristic classifier based on regular expressions. The classifier comprises eleven rules, each targeting partially distinct types of surface-level property (see Table~\ref{tab:ling-rules}). A hypothesis was classified as linguistic if it matched \emph{at least one} rule. Multiple rules could be triggered on the same hypothesis: for example, the hypothesis ``The items all have two-word common names and start with the same letter'' simultaneously matches \textsc{letter}, \textsc{word\_count}, \textsc{naming}, and \textsc{meta\_linguistic\_structure}. Consequently, per-rule counts reflect the number of linguistic-pattern matches across all hypotheses (with overlaps permitted), not the number of distinct games or turns. Figure~\ref{fig:rule_count} shows the per-rule counts across all game turns.

\begin{figure*}[t!]
    \centering
    \begin{subfigure}[t]{0.48\linewidth}
    \centering
    \includegraphics[width=\linewidth]{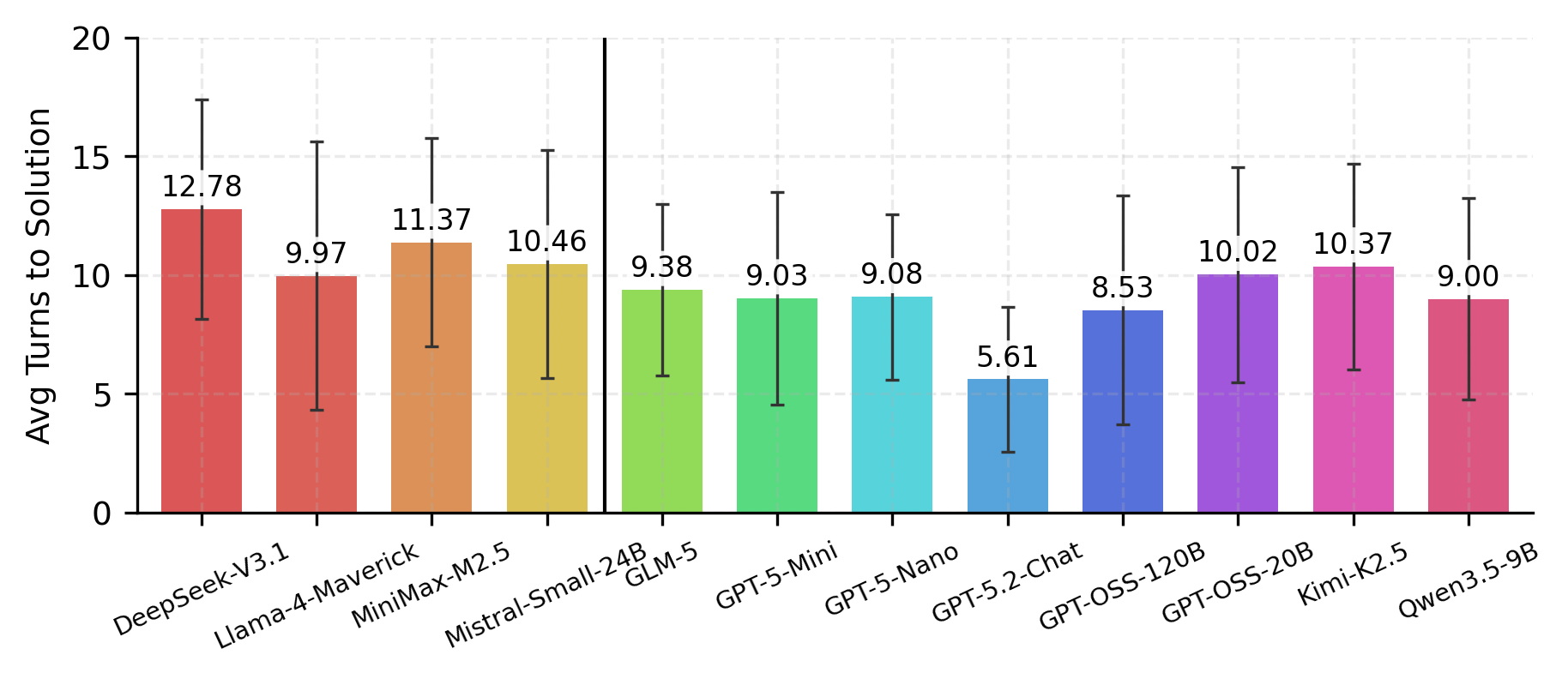}
    \end{subfigure}
    \hfill
    \begin{subfigure}[t]{0.48\linewidth}
    \centering
    \includegraphics[width=\linewidth]{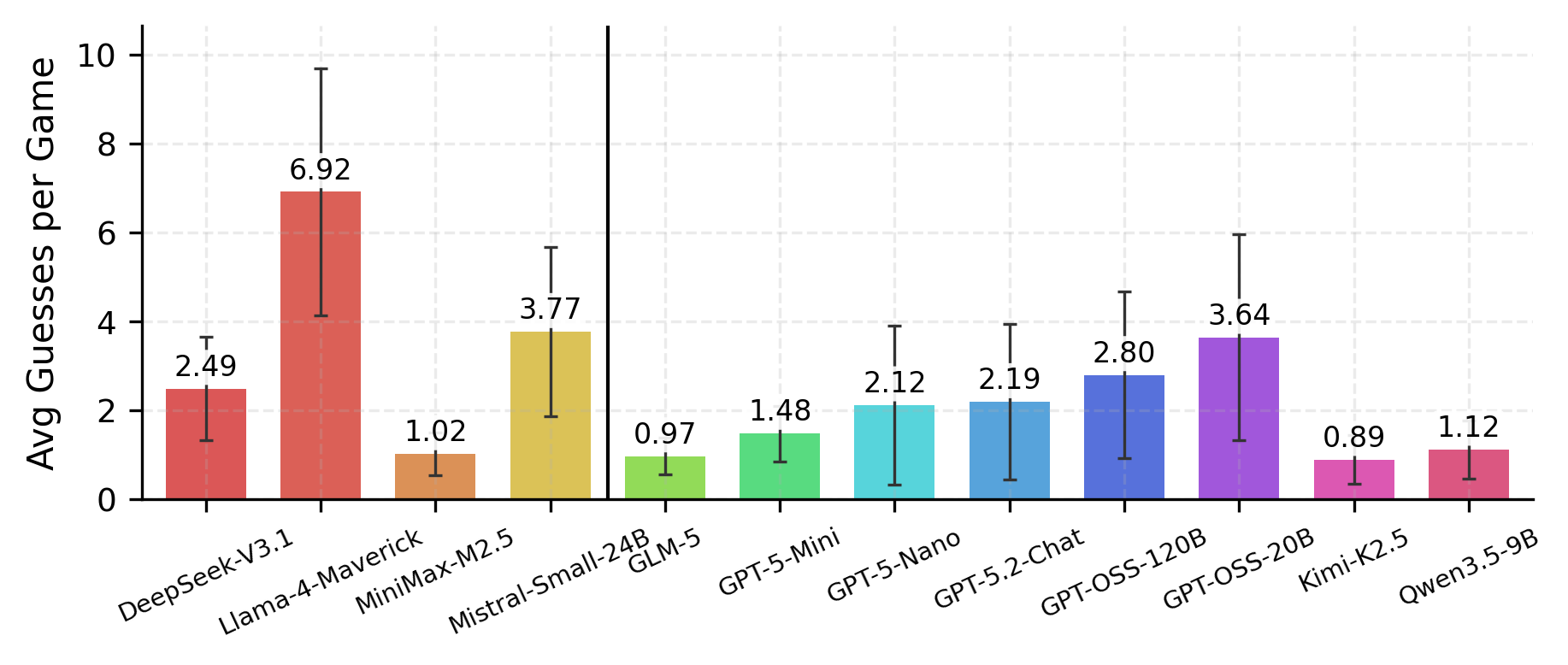}
    \end{subfigure}
    \caption{\textbf{Left:} The average number of turns required to correctly identify the target rule $R$, computed over successful games only. \textbf{Right:} The average number of \texttt{Guess} actions performed per game. In both plots, models to the left of the vertical black line are \textbf{instruction-tuned} variants; models to the right are \textbf{reasoning} variants.}
    \label{fig:guess_turns}
\end{figure*}

\section{Additional Results}
\label{app:results}

\subsection{Additional Player Metrics}

In addition to the player metrics presented in the main paper, we also show here two additional metrics that are informative for understanding the overall quality of players.

\paragraph{Average Turns to Solution.} The average number of turns required to correctly identify the target rule $R$, computed over successful games only.

\paragraph{Average Guesses per Game.} The average number of \texttt{Guess} actions submitted per game. While guessing can reveal that a hypothesis is wrong by receiving the string ``Incorrect'' as feedback, it reflects a passive strategy compared to targeted \texttt{Test} actions designed to probe the boundaries of the current hypothesis.

\vspace{1mm}

Figure~\ref{fig:guess_turns} reports the results for each model on these two additional metrics. In terms of turn efficiency, most models fell in the range between 8 and 11 turns, and we did not observe as large a gap between reasoning and instruction-tuned models as we did for other metrics, such as success rates. A notable positive exception is GPT-5.2-Chat, which stood out as the most efficient model overall, solving games in just 5.61 turns on average. Regarding the average guesses per game, Llama-4-Maverick stood out as the model with the most passive strategy, combining one of the highest positive testing bias scores (81.8\%, see Figure~\ref{fig:results_main}) with the highest average number of guesses per game (6.92), suggesting it relied on repeated guessing rather than active falsification to make progress. At the other extreme, the two largest open models, GLM-5 and Kimi-K2.5, showed the lowest guessing rates (GLM-5: 0.97, Kimi-K2.5: 0.89), indicating that they did not commit to hypotheses prematurely.

\begin{figure*}[h]
    \centering
    \begin{subfigure}[t]{0.48\linewidth}
        \centering
        \includegraphics[width=\linewidth]{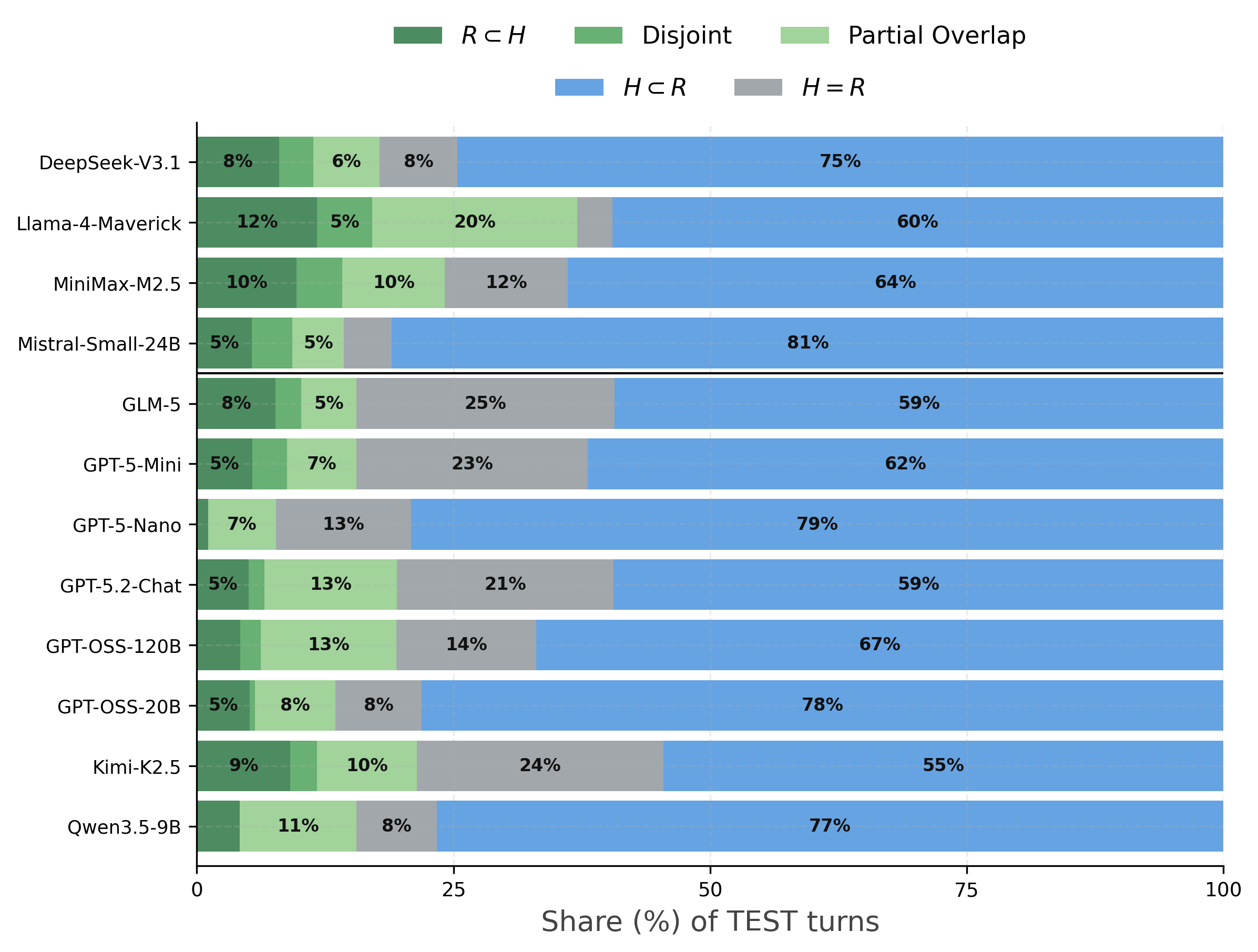}
    \end{subfigure}
    \hfill
      \begin{subfigure}[t]{0.48\linewidth}
        \centering
        \includegraphics[width=\linewidth]{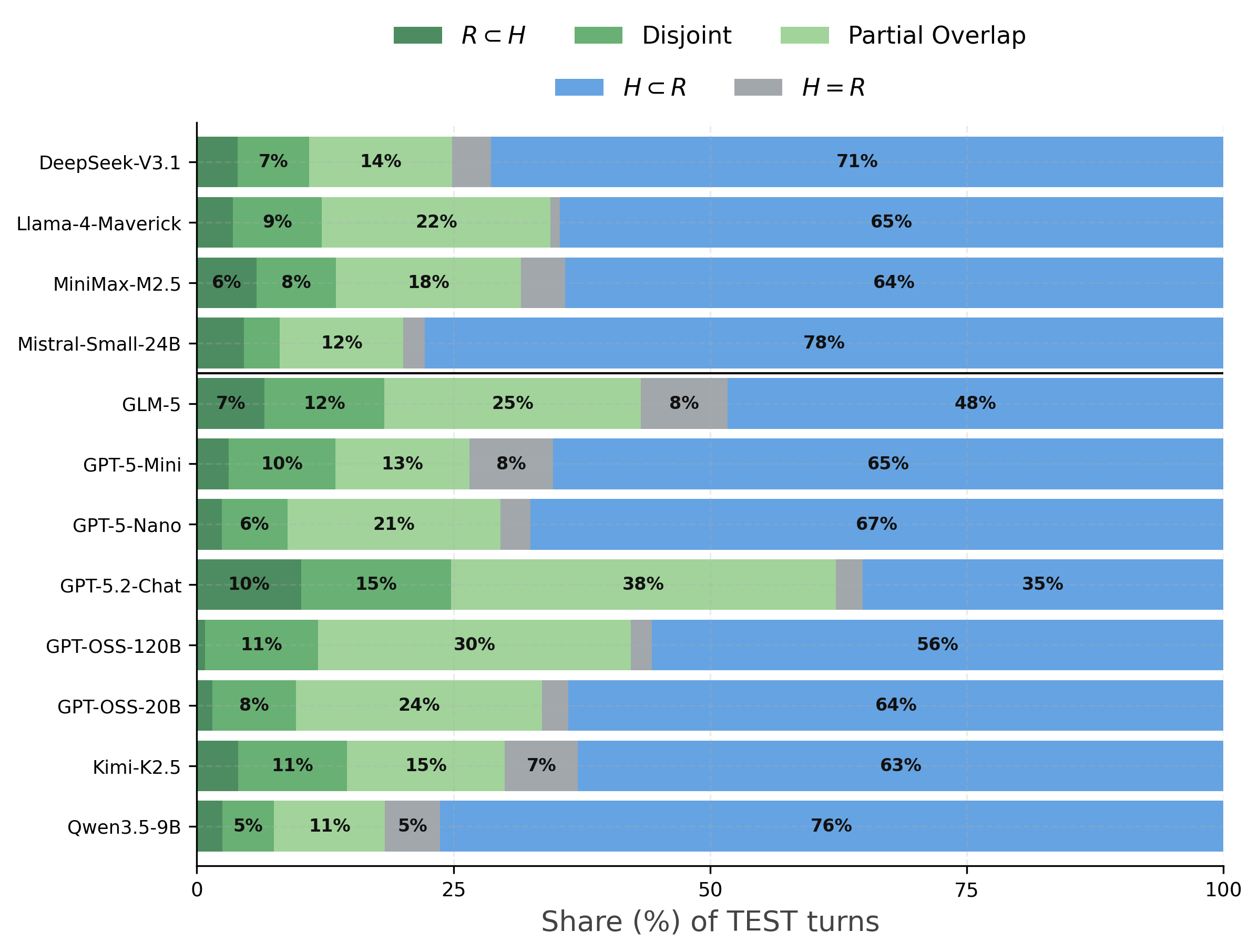}
    \end{subfigure}
    \caption{Distribution of relations between $H$ and $R$ across \texttt{Test} turns, split by outcome. \textbf{Left:} Successful games only. \textbf{Right:} Failed games only. Successful games are dominated by the $H \subset R$ configuration (blue) reflecting coherent upward revision. Failed games show a substantially higher proportion of Partial Overlap (green) and Disjoint (yellow) turns across most models, indicating that failure is most commonly associated with failure in exploring the hierarchical structure of the hypothesis space.}
    \label{fig:relation_success_fail}
\end{figure*}

\subsection{Relation Distribution in Successful and Failed Games}
\label{app:error}

Figures~\ref{fig:relation_success_fail} shows the distribution of relations between $H$ and $R$ across \texttt{Test} turns, split by whether the game was ultimately successful or not. In both conditions, $H \subset R$ (Hypothesis in Target) was the dominant configuration, confirming that models generally maintained hypotheses that were too narrow throughout the game. However, noticeable differences emerged between the two groups.

In successful games, the distribution was more concentrated around $H \subset R$, suggesting that successful play tended to follow a clean upward revision trajectory that mirrored the structure of the hierarchical hypothesis space derived from the WordNet taxonomy.

In failed games, the partial overlap configuration was the second most common relation for most models, particularly among stronger models: GPT-5.2-Chat reached nearly 40\% Partial Overlap turns, while GPT-OSS-120B and GLM-5 reached approximately 30\% and 25\%, respectively. This suggests that, in failed games, these models frequently arrived at hypotheses similar to $R$ but failed to follow a clear upward path. The proportion of disjoint cases also tended to be higher in failed games, though to a lesser extent, suggesting that fully off-track hypotheses were associated with failure. Finally, stronger models also exhibited cases in which they had effectively reached the target rule but failed to commit to a \texttt{Guess}, either abandoning it prematurely or exhausting the turn limit.

\subsection{Decomposition by target rule}

\begin{figure*}[h]
    \centering
    \includegraphics[width=\linewidth]{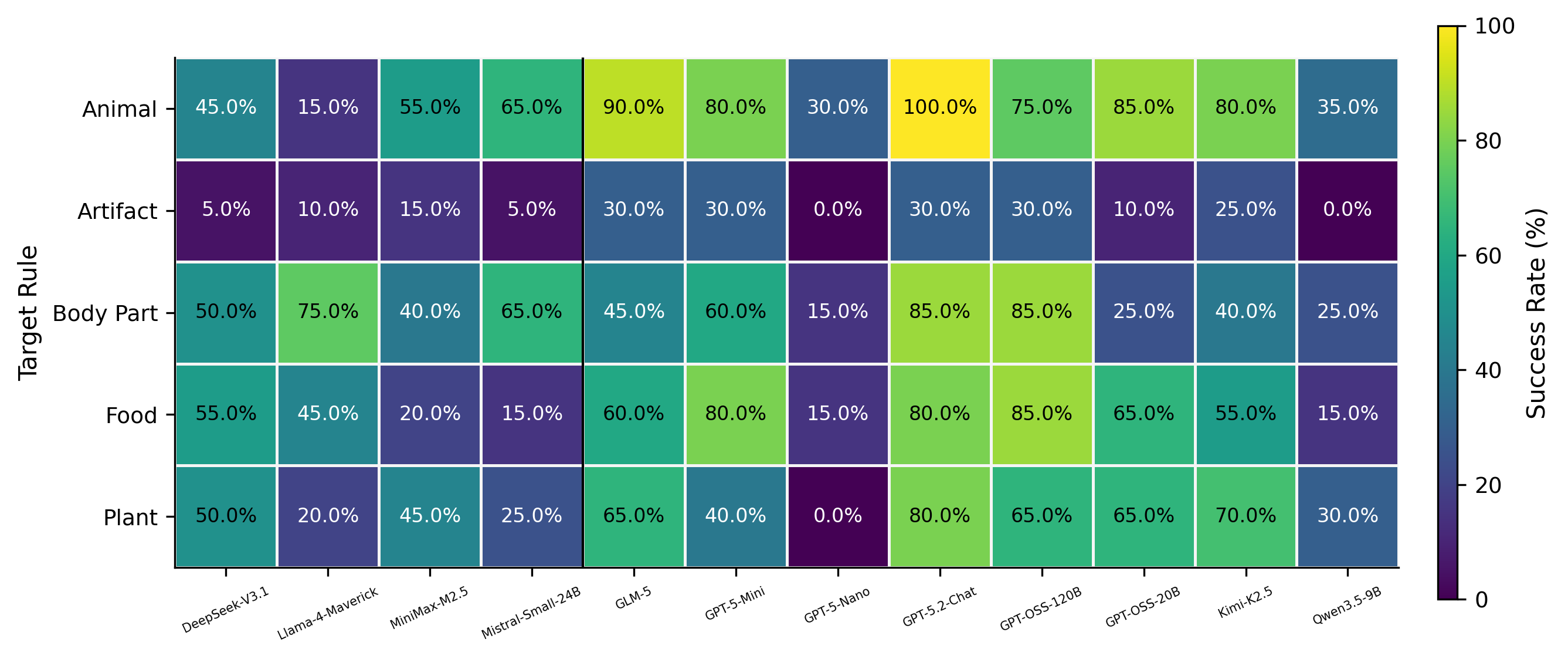}
    \\[6pt]
    \includegraphics[width=\linewidth]{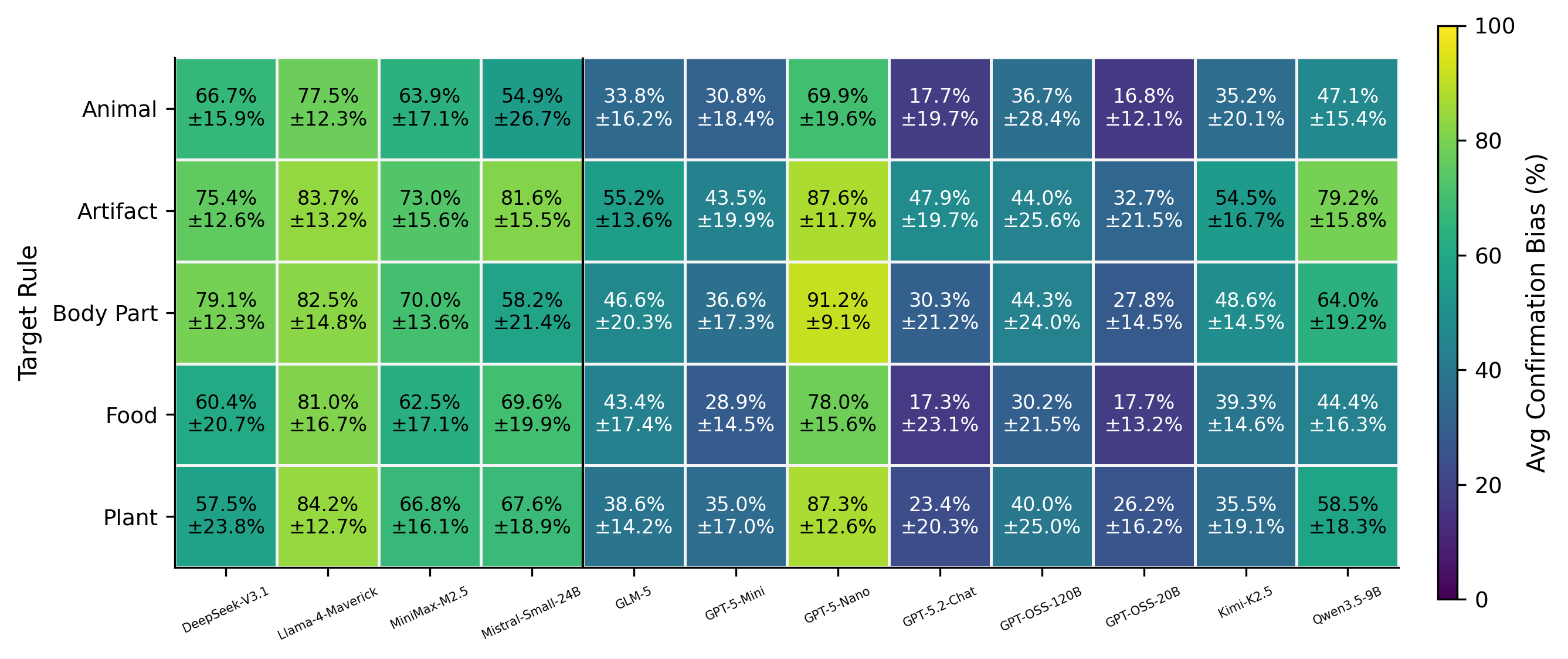}
    \\[6pt]
    \includegraphics[width=\linewidth]{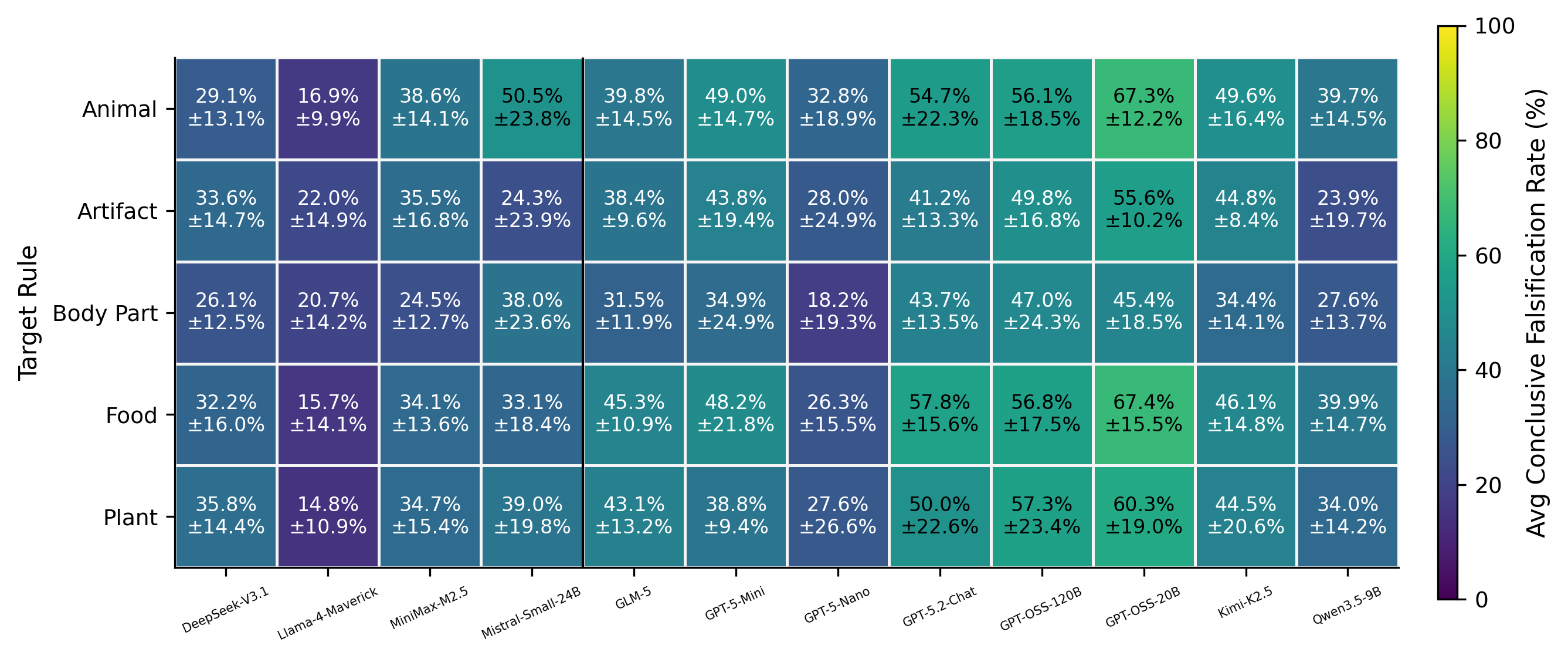}
    \caption{\textbf{Top:} success rate. \textbf{Center:} average Confirmation bias. \textbf{Bottom:} average conclusive falsification rate. All metrics are decomposed by target rule.}
    \label{fig:by_target_one}
\end{figure*}

\begin{figure*}[h]
    \centering
    \includegraphics[width=\linewidth]{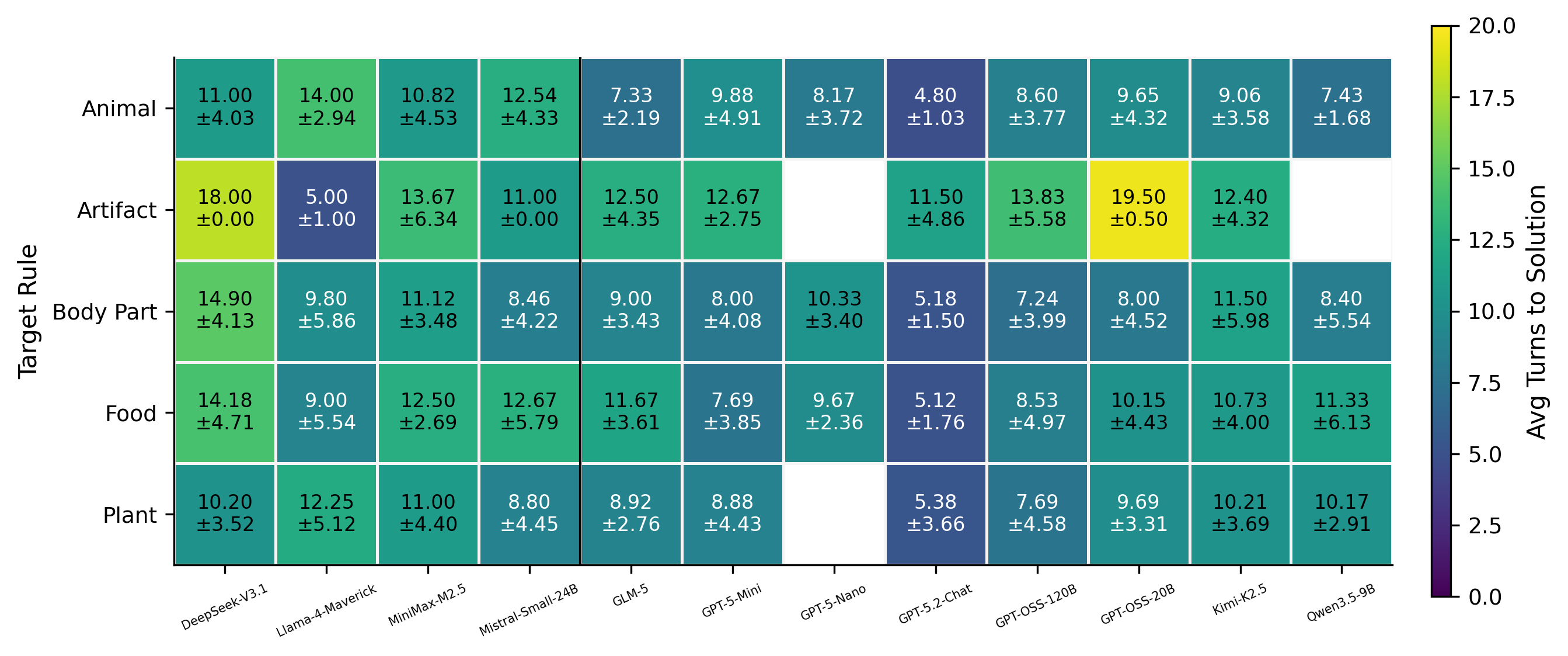}
    \\[6pt]
    \includegraphics[width=\linewidth]{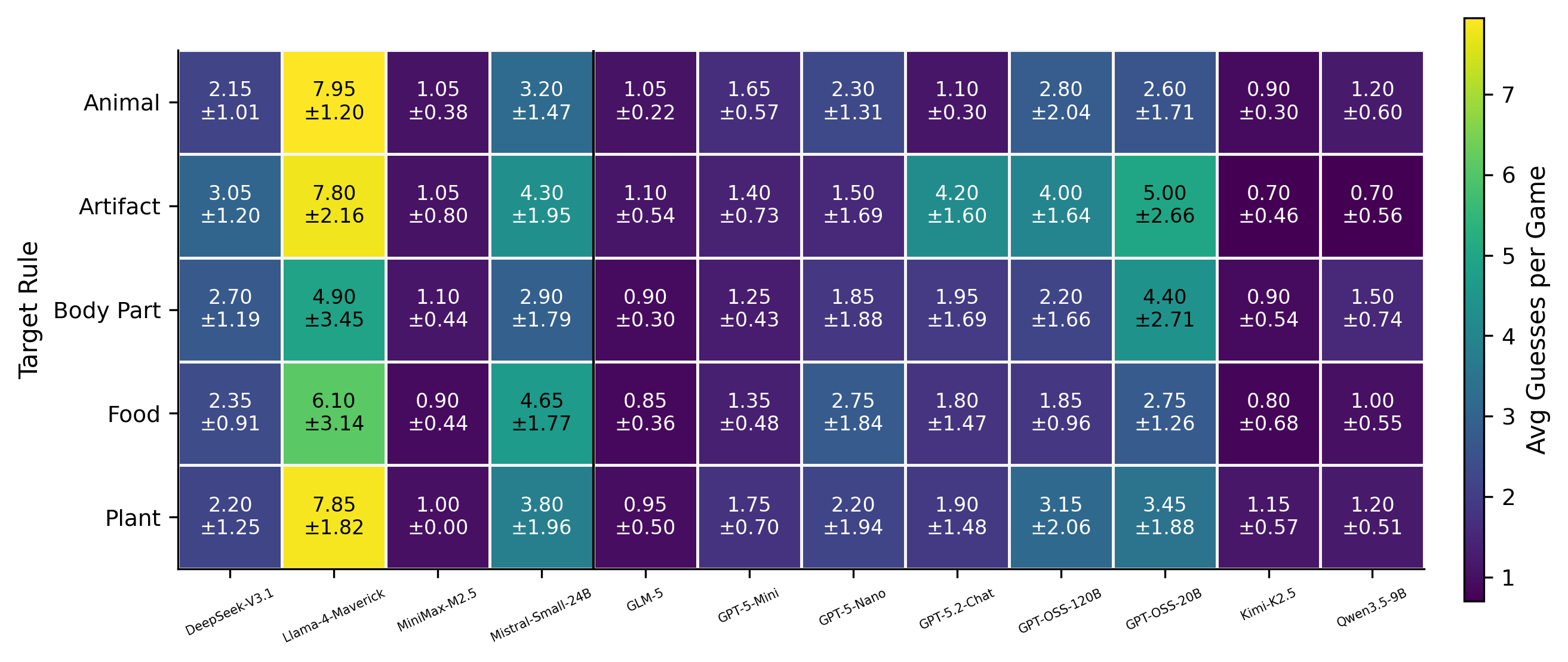}
    \caption{\textbf{Top:} average turns to solution. \textbf{Bottom:} average guesses per game. All metrics are decomposed by target rule. Empty cells in the turns heatmap indicate zero successful games for that model-target rule combination.}
    \label{fig:by_target_two}
\end{figure*}

Figures~\ref{fig:by_target_one} and \ref{fig:by_target_two} report success rate, Confirmation bias, conclusive falsification rate, average turns to solution, and average guesses per game decomposed by target rule ($R \in$ \{``animal'', ``artifact'', ``body part'', ``food'', ``plant''\}). The most consistent finding is that ``artifact'' was by far the hardest target rule across all models and metrics: success rates were uniformly low (0--30\% for most models), Confirmation bias was highest, and conclusive falsification rates were lowest. This was likely because artifact subcategories spanned a semantically heterogeneous space, making it harder to form hypotheses that generalized cleanly upward through the taxonomy. By contrast, ``animal'' was the easiest target for most reasoning models, with GPT-5.2-Chat achieving a 100\% success rate and several others exceeding 75\%.

Regarding average guesses per game, Llama-4-Maverick submitted an average of nearly 8 guesses across all target properties, far exceeding every other model and confirming that its strategy relied heavily on passive guessing rather than active evidence gathering regardless of the semantic domain. In the artifact category, well-performing models also showed elevated guessing behavior, suggesting that the semantic difficulty of this category induced more premature commitment across models. Turn efficiency broadly mirrored success rate patterns, with \textit{animal} games being solved fastest by reasoning models and \textit{artifact} games either taking the most turns or remaining unsolved within the turn limit, as indicated by empty cells in the turns heatmap for models with zero success in that category.

\subsection{Decomposition by Outcome}

Figure~\ref{fig:by_outcome} reports the average guesses per game and Confirmation bias split by game outcome (Succeeded vs.\ Failed).

In terms of average guesses (left panel), failed games involved substantially more guesses than successful ones across all models. The effect was most pronounced for Llama-4-Maverick, which averaged 8.5 guesses per failed game compared to 3.8 in successful ones, and for GPT-5.2-Chat, GPT-OSS-120B, and GPT-OSS-20B, which all approached 4--5 guesses per failed game while remaining below 2 in successful ones. This suggests that models relied on guessing and ``Incorrect'' feedback to disconfirm hypotheses when they were not able to efficiently explore the hypothesis space.

Similarly, regarding Confirmation bias (right panel), failed games exhibited higher Confirmation bias than successful ones for nearly every model. The gap was largest for stronger reasoning models: GPT-5.2-Chat showed 20\% bias in successful games versus 50\% in failed ones, while GPT-OSS-20B showed 17\% versus 53\%. This provides further within-model evidence that negative testing was associated with success and represented the most efficient strategy for reaching the target.

\subsection{Qualitative Game Examples}
\label{app:game_examples}
 
To give the reader a better understanding of the game dynamics, we present one successful (Table \ref{tab:game1}) and one failed (Table \ref{tab:game2}) complete game trace produced by GPT-5.2-Chat. In each trace the player's strategy in \texttt{Test} turns is labeled with \textit{positive} for positive tests and \textit{negative} for negative tests.

\section{Use of Generative AI Tools}
We used GitHub Copilot and Claude Code as assistant tools for parts of the project’s source code development, while ChatGPT was used to correct minor grammatical errors in the manuscript.

\begin{figure*}[t]
    \centering
    \begin{minipage}{0.48\linewidth}
        \centering
        \includegraphics[width=\linewidth]{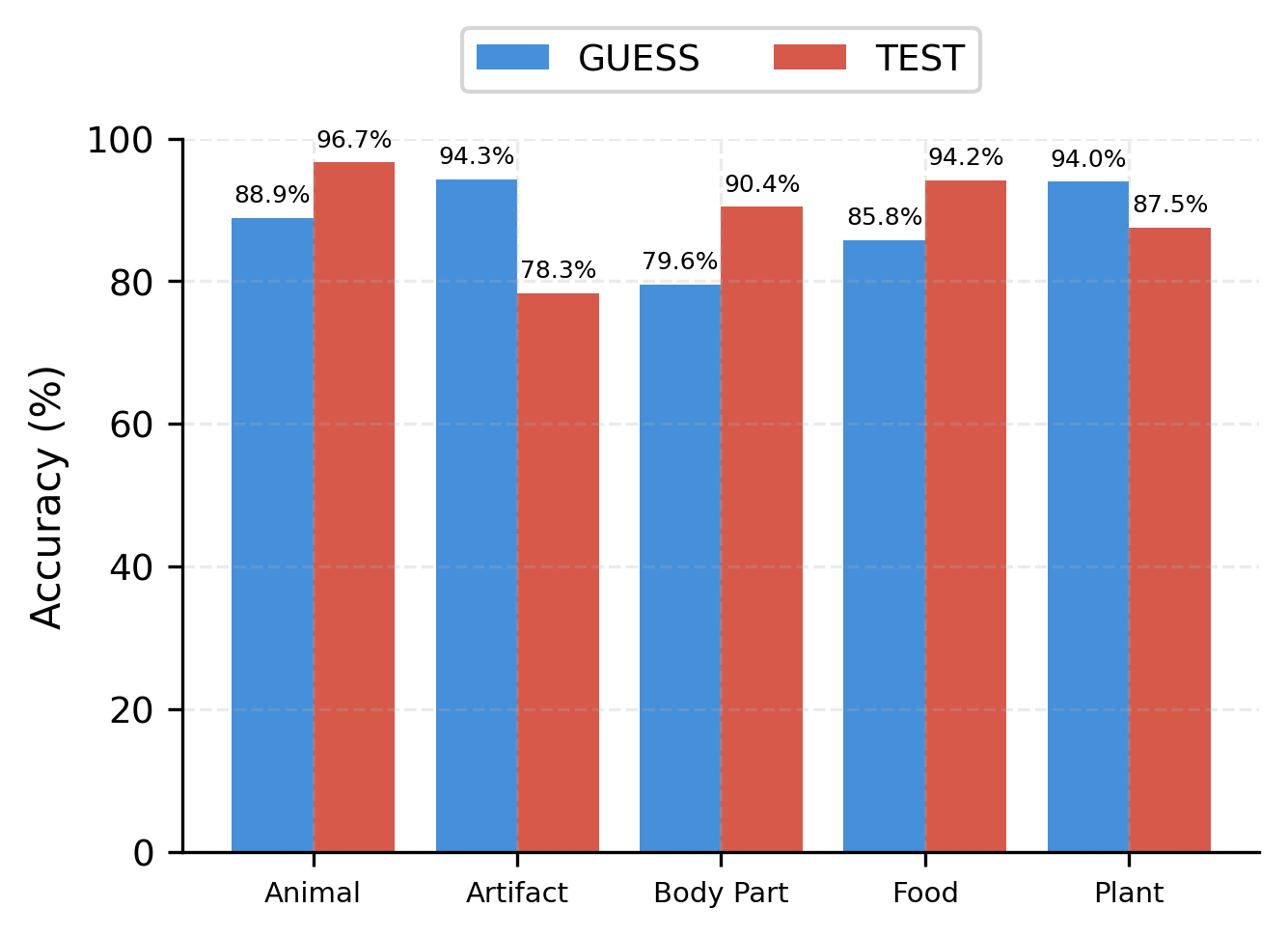}
    \end{minipage}
    \hfill
    \begin{minipage}{0.48\linewidth}
        \centering
        \includegraphics[width=\linewidth]{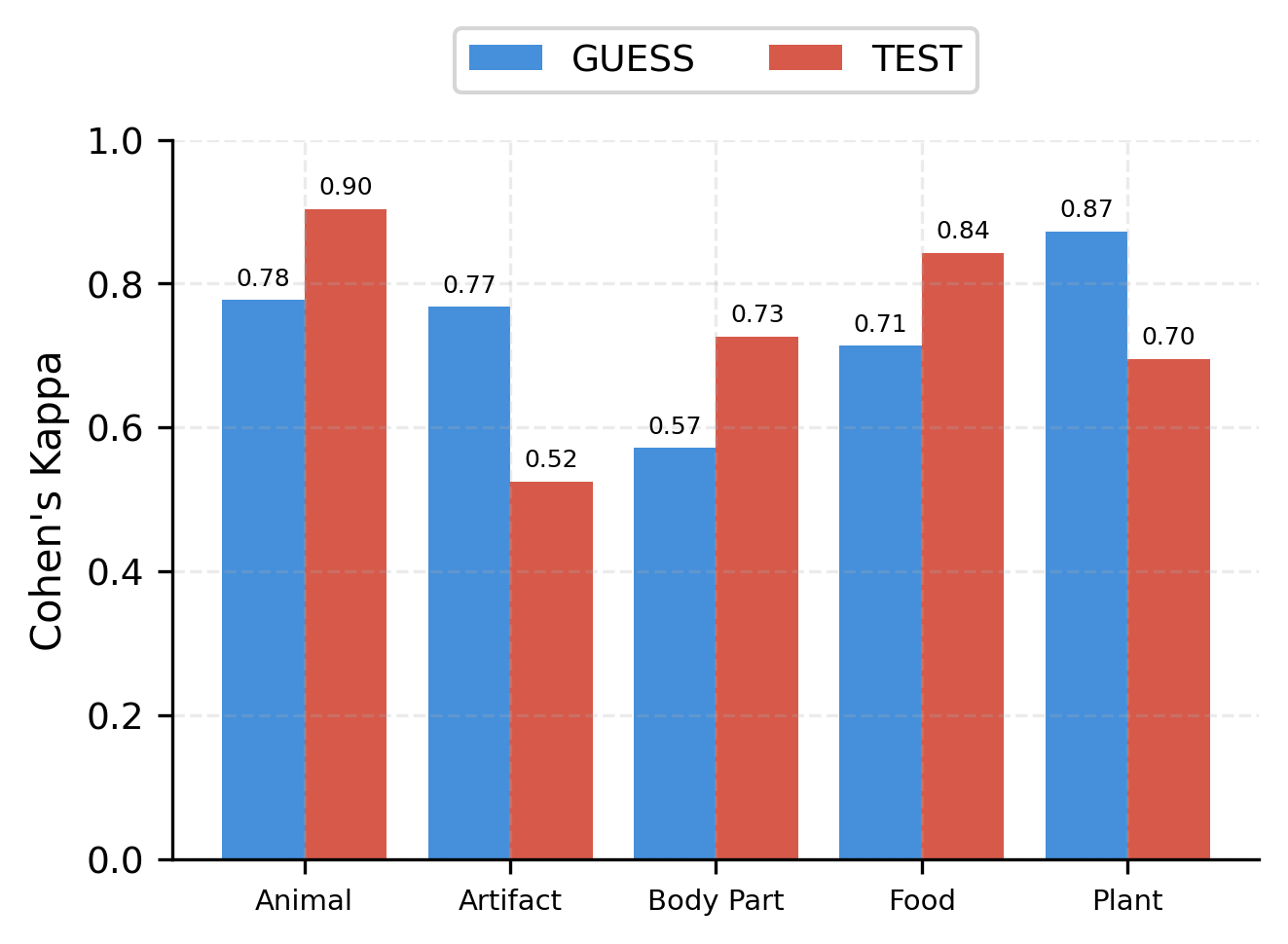}
    \end{minipage}
    \caption{Oracle accuracy and agreement with human annotator decomposed by target rule. \textbf{Left:} Raw accuracy. \textbf{Right:} Cohen's $\kappa$ agreement.}
    \label{fig:by_outcome}
\end{figure*}

\begin{figure*}[t]
    \centering
    \begin{minipage}{0.48\linewidth}
        \centering
        \includegraphics[width=\linewidth]{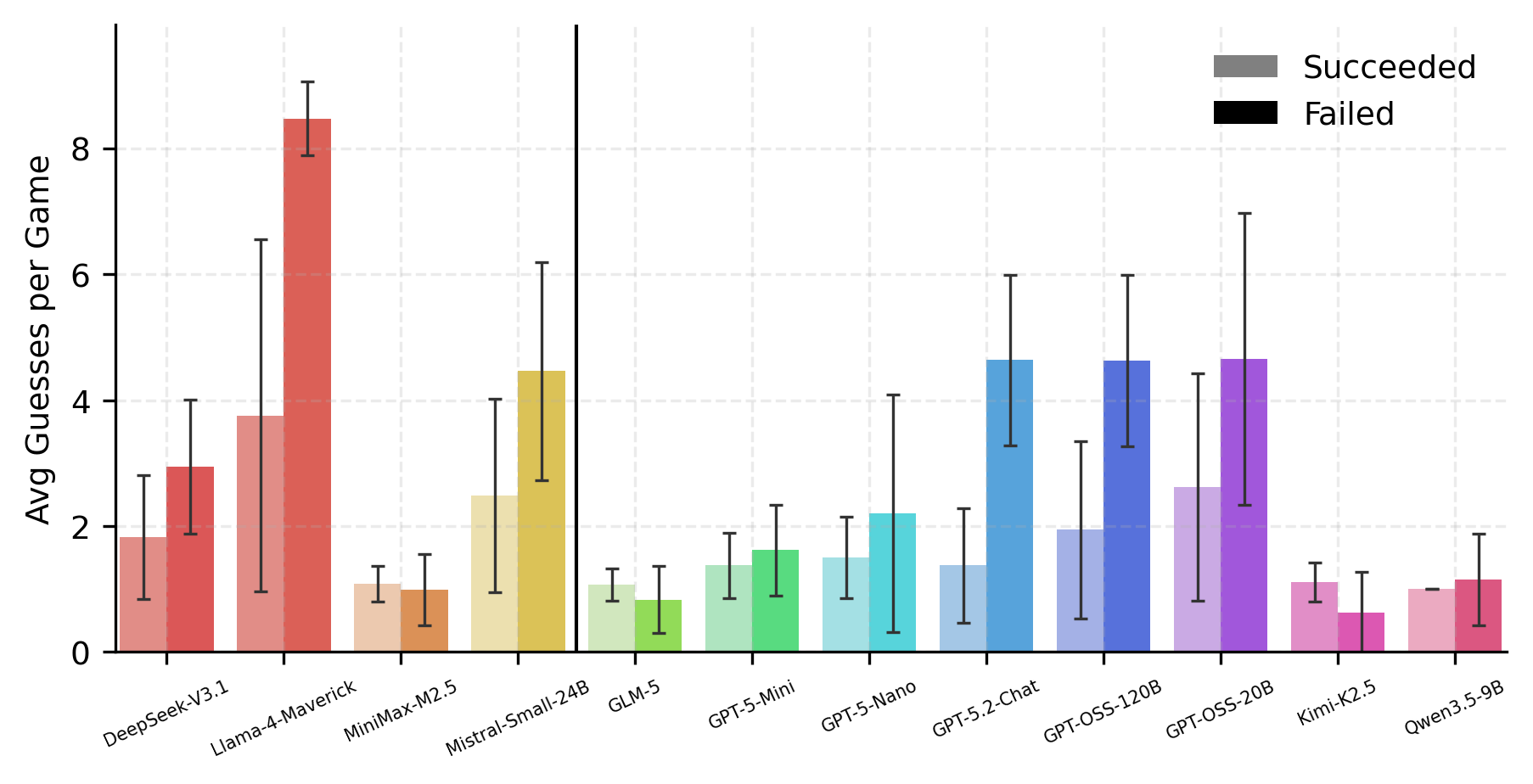}
    \end{minipage}
    \hfill
    \begin{minipage}{0.48\linewidth}
        \centering
        \includegraphics[width=\linewidth]{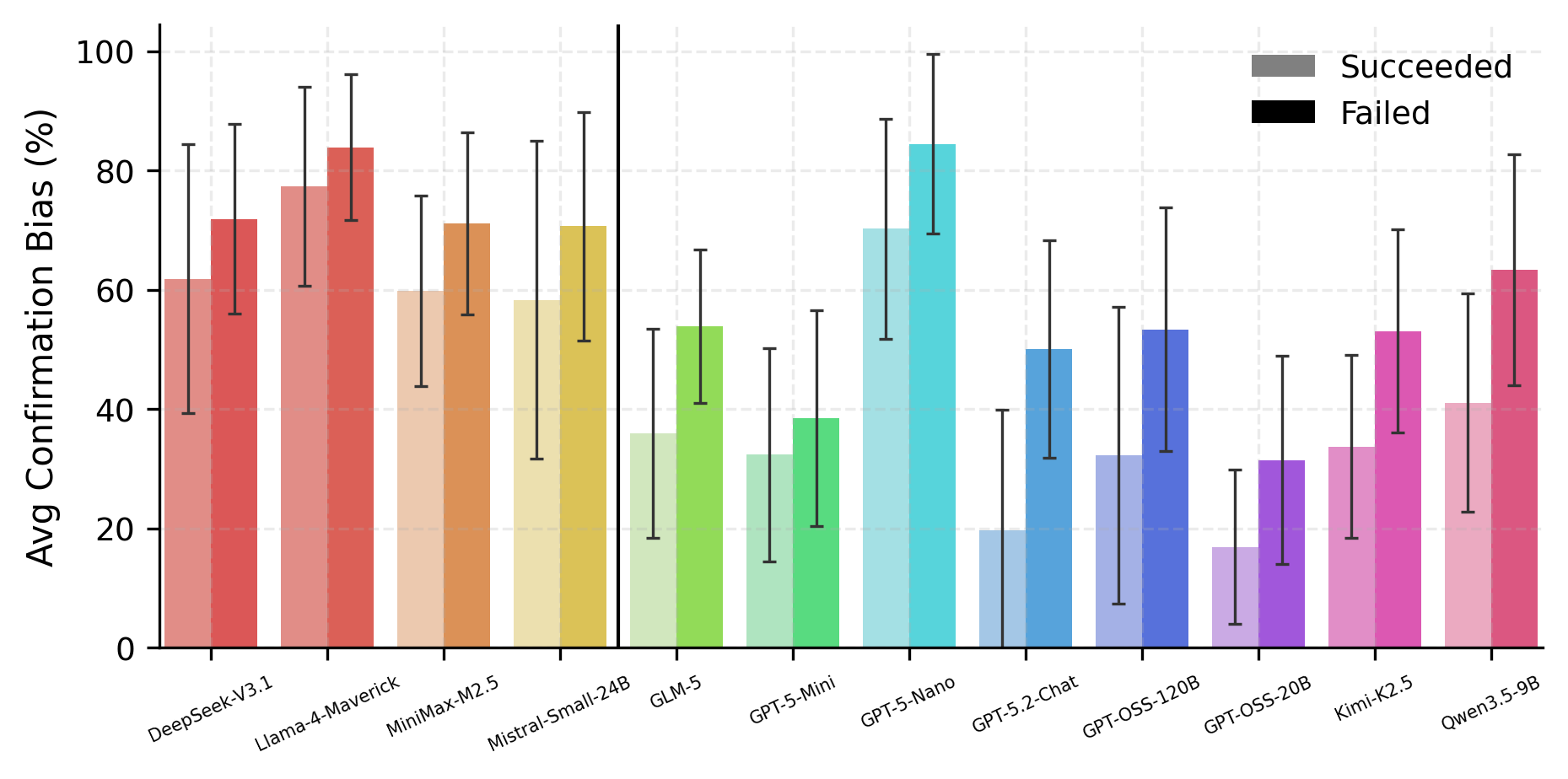}
    \end{minipage}
    \caption{Results decomposed by game outcome (Succeeded vs.\ Failed). \textbf{Left:} Average 
    guesses per game. \textbf{Right:} Average positive testing bias. Light bars correspond to 
    successful games; dark bars to failed games. Error bars show standard deviation.}
    \label{fig:by_outcome}
\end{figure*}

% -- PLAYER PROMPTS (full-width) -------------------------------------------
\begin{table*}[ht]

% ---------- Box 1 ----------

\begin{tcolorbox}[
  colback=white,
  colframe=black,
  coltitle=white,
  rounded corners,
  title={\sffamily Player System Prompt},
]
{\small\ttfamily
You are playing a property discovery game. You are given 3 items that
share a common property. Your goal is to guess the property by testing
new items.
 
\smallskip
You have a maximum of \{\{max\_turns\}\} turns.
 
\smallskip
What to expect from the oracle:\newline
- After a \texttt{Test} action, the oracle will reply with: ``Oracle's judgment: CONFORM''
  or ``Oracle's judgment: DO NOT CONFORM''.\newline
- After a \texttt{Guess} action, the oracle will reply with: ``Oracle's judgment: INCORRECT''
  if you're wrong; if you're correct the game ends.
 
\smallskip
In each turn, you can perform one of two actions:\newline
\hspace*{1em}1.\ Propose 3 new items to test your current hypothesis.\newline
\hspace*{1em}2.\ Make a guess of the property if you think you know it.
 
\smallskip
The purpose of testing is to gather information to refine your hypothesis.
Once you believe you have enough information, you can make a guess.
If your guess is correct, you win. If it's incorrect, you can keep
testing and guessing until you find the right property or run out of turns.
 
\smallskip
Your response must be valid JSON in one of these formats:
 
\smallskip
\textit{For testing:}\newline
\{\newline
\hspace*{2em}"action": "test",\newline
\hspace*{2em}"items": ["item1", "item2", "item3"],\newline
\hspace*{2em}"hypothesis": "your current hypothesis",\newline
\hspace*{2em}"reasoning": "explain your thinking"\newline
\}
 
\smallskip
\textit{For guessing:}\newline
\{\newline
\hspace*{2em}"action": "guess",\newline
\hspace*{2em}"property": "your final guess as a single word or phrase",\newline
\hspace*{2em}"reasoning": "explain why you're confident"\newline
\}
}
\end{tcolorbox}
 
% ---------- Box 2 ----------

\begin{tcolorbox}[
  colback=white,
  colframe=black,
  coltitle=white,
  rounded corners,
  title={\sffamily Player's first user message},
]
{\small\ttfamily
Initial examples that share a common property:\newline
\hspace*{1em}1.\ \{\{initial\_examples[0]\}\}\newline
\hspace*{1em}2.\ \{\{initial\_examples[1]\}\}\newline
\hspace*{1em}3.\ \{\{initial\_examples[2]\}\}
 
\smallskip
What is your first move? Respond with JSON only.
}
\end{tcolorbox}

\caption{System prompt and first user role interaction provided to the player. The system prompt specifies the rules of the game and is provided once at the start of each game; the initial message introduces the three initial examples. Double curly-brace placeholders are filled before each game.}
\label{box:player}
\end{table*}
 
% -- ORACLE PROMPTS (full-width) --------------------------------------------

\begin{table*}[ht]

% ---------- Box 1 ----------
\begin{tcolorbox}[
  colback=white,
  colframe=black,
  coltitle=white,
  rounded corners,
  title={{\sffamily Judge \texttt{Test} triples against target rule}},
]
{\small\ttfamily
You are an oracle in a property discovery game.\newline
The target rule is: ``\{\{target\_property\}\}''
 
\smallskip
The player has proposed these items:\newline
\hspace*{1em}1.\ \{\{items[0]\}\}\newline
\hspace*{1em}2.\ \{\{items[1]\}\}\newline
\hspace*{1em}3.\ \{\{items[2]\}\}
 
\smallskip
Do ALL three items conform to the property ``\{\{target\_property\}\}''?
 
\smallskip
Respond with ONLY a JSON object:\newline
\{\newline
\hspace*{2em}"conforms": true or false,\newline
\hspace*{2em}"explanation": "brief explanation of your judgment"\newline
\}
}

\end{tcolorbox}

% ---------- Box 2 ----------
\begin{tcolorbox}[
  colback=white,
  colframe=black,
  coltitle=white,
  rounded corners,
  title={{\sffamily Classify player's test strategy}},
]
{\small\ttfamily
You are an oracle in a property discovery game.\newline
The property to evaluate is: ``\{\{property\_hypothesis\}\}''
 
\smallskip
The player has proposed these items:\newline
\hspace*{1em}1.\ \{\{items[0]\}\}\newline
\hspace*{1em}2.\ \{\{items[1]\}\}\newline
\hspace*{1em}3.\ \{\{items[2]\}\}
 
\smallskip
Player's current hypothesis:\newline
``\{\{property\_hypothesis\}\}''
 
\smallskip
Player's reasoning for choosing these items:\newline
``\{\{player\_reasoning\}\}''
 
\smallskip
Tasks:\newline
1) Infer the player's *intention* for this \texttt{Test} based on their hypothesis and reasoning.\newline
\hspace*{2em}- "confirm" = they expect all three items to conform to their hypothesis\newline
\hspace*{2em}- "falsify" = they expect at least one item to NOT conform to their hypothesis\newline
2) Separately, judge whether all three items actually conform to the hypothesis.
 
\smallskip
Respond with ONLY a JSON object:\newline
\{\newline
\hspace*{2em}"intention": "confirm" or "falsify",\newline
\hspace*{2em}"expected\_conforms": true or false,\newline
\hspace*{2em}"conforms": true or false,\newline
\hspace*{2em}"explanation": "brief explanation"\newline
\}
}

\end{tcolorbox}

% ---------- Box 3 ----------
\begin{tcolorbox}[
  colback=white,
  colframe=black,
  coltitle=white,
  rounded corners,
  title={{\sffamily Judge \texttt{Guess} against the target rule}},
]
{\small\ttfamily
You are an oracle in a property discovery game.\newline
The target rule is: ``\{\{target\_property\}\}''
 
\smallskip
The player has guessed: ``\{\{guessed\_property\}\}''
 
\smallskip
Are these semantically equivalent (same meaning, possibly different wording)?
 
\smallskip
Respond with ONLY a JSON object:\newline
\{\newline
\hspace*{2em}"correct": true or false,\newline
\hspace*{2em}"explanation": "brief explanation"\newline
\}
}
\end{tcolorbox}

\caption{Complete prompts provided to the oracle. Each prompt specifies a different type of annotation that the oracle has to provide. Double curly-brace placeholders are filled before each game.}
\label{box:oracle}

\end{table*}

% -- ANNOTATION PROMPT (full-width) -------------------------------------------
\begin{table*}[ht]
\begin{tcolorbox}[
  colback=white,
  colframe=black,
  coltitle=white,
  rounded corners,
  title={\sffamily Annotation Prompt},
]
{\small\ttfamily
You are annotating saved game turns from an inductive reasoning experiment.

\smallskip
Treat the target rule and the current hypothesis as semantic sets or classes of entities.
Decide the relation between those two sets using exactly one label from this closed set:

\smallskip
- identical: the two sets are the same.\newline
- disjoint: the two sets share no members.\newline
- partial\_overlap: the two sets overlap, but neither includes the other.\newline
- hypothesis\_included\_in\_target: the hypothesis set is a proper subset of the target set.\newline
- target\_included\_in\_hypothesis: the target set is a proper subset of the hypothesis set.

\smallskip
Interpret terms by their normal semantic meaning, not by string overlap.
If one label is more specific than another, choose the more specific label:\newline
- Prefer identical over either inclusion label when the sets are the same.\newline
- Prefer an inclusion label over partial\_overlap when one set fully contains the other.

\smallskip
Examples using target rule ``animal'':\newline
1.\ Target: ``animal'' | Hypothesis: ``animal'' -> identical\newline
2.\ Target: ``animal'' | Hypothesis: ``mammal'' -> hypothesis\_included\_in\_target\newline
3.\ Target: ``animal'' | Hypothesis: ``living thing'' -> target\_included\_in\_hypothesis\newline
4.\ Target: ``animal'' | Hypothesis: ``carnivore'' -> partial\_overlap\newline
5.\ Target: ``animal'' | Hypothesis: ``mineral'' -> disjoint

\smallskip
Now annotate this case.\newline
target rule: ``\{\{target\_property\}\}''\newline
Current hypothesis: ``\{\{hypothesis\}\}''

\smallskip
Respond with JSON only in this format:

\smallskip
\{\newline
\hspace*{2em}"relation": "one of identical, disjoint, partial\_overlap, hypothesis\_included\_in\_target, target\_included\_in\_hypothesis",\newline
\hspace*{2em}"explanation": "at most 12 words"\newline
\}

\smallskip
Do not add any extra keys, markdown, or prose outside the JSON object.
}

\end{tcolorbox}
\caption{System prompt used for semantic set-relation annotation. The annotator classifies the relationship between a target rule and a hypothesis using a fixed label set. An explanation is included to better understand the provided annotation. Double curly-brace placeholders are filled before each API call.}
\label{box:annotation}
\end{table*}

\begin{table*}[ht]
\begin{tcolorbox}[
  colback=white,
  colframe=successcolor,
  coltitle=white,
  rounded corners,
  title={\sffamily Successful Game}
]
\scriptsize\setlength{\parskip}{1pt}

\textbf{target rule (hidden to player):} body part \\
\textbf{Sampling Category (hidden to player):} sense organ \\
\textbf{Initial Items:} thermoreceptor, lateral line, third eye

\smallskip\hrule\smallskip
 
\textbf{T1} \textit{(positive)}\enspace
\textbf{Items:} electroreceptor, Jacobson's organ, ampullae of Lorenzini\enspace
\textbf{H:} Specialized sensory organs/receptors that detect environmental stimuli.\enspace
\textbf{R:} Similar to the initial examples; testing other exotic sensory structures.\enspace
$\to$ \texttt{CONFORM}

\smallskip\hrule\smallskip
  
\textbf{T2} \textit{(negative)}\enspace
\textbf{Items:} retina, cochlea, neuron\enspace
\textbf{H:} Specialized anatomical structures dedicated to detecting specific sensory stimuli.\enspace
\textbf{R:} Including a neuron---a general nerve cell, not a sense organ---to probe the property's boundary.\enspace
$\to$ \texttt{DO NOT CONFORM}
 
\smallskip\hrule\smallskip
 
\textbf{T3} \textit{(negative)}\enspace
\textbf{Items:} magnetoreceptor, infrared pit organ, taste bud\enspace
\textbf{H:} Specialized sensory organs providing non-classical senses beyond the standard five.\enspace
\textbf{R:} Testing whether organs tied to well-known senses (taste bud) also conform.\enspace
$\to$ \texttt{CONFORM}
 
\smallskip\hrule\smallskip
 
\textbf{T4} \textit{(guess)}\enspace
\textbf{Property:} ``sensory organs or sensory receptors''\enspace
\textbf{R:} All tested sensory structures conform; neuron does not.\enspace
$\to$ \texttt{INCORRECT}
 
\smallskip\hrule\smallskip
 
\textbf{T5} \textit{(negative)}\enspace
\textbf{Items:} retina, cochlea, olfactory epithelium\enspace
\textbf{H:} Sensory structures that are not the primary classical human sense organs.\enspace
\textbf{R:} Testing canonical sensory organs to see if they conform despite belonging to the five standard senses.\enspace
$\to$ \texttt{CONFORM}
 
\smallskip\hrule\smallskip
 
\textbf{T6} \textit{(negative)}\enspace
\textbf{Items:} heart, kidney, lung\enspace
\textbf{H:} Anatomical structures specifically involved in sensory perception.\enspace
\textbf{R:} Testing organs with purely physiological (non-sensory) roles to check whether the property is broader than sensory function.\enspace
$\to$ \texttt{CONFORM}
 
\smallskip\hrule\smallskip
 
\textbf{T7} \textit{(guess)}\enspace
\textbf{Property:} ``anatomical body parts''\enspace
\textbf{R:} Non-sensory organs (heart, kidney, lung) still conform; the property must cover all body parts.\enspace
$\to$ \texttt{CORRECT}

\smallskip\hrule\smallskip
\textcolor{successcolor}{\textbf{SUCCESS \checkmark}}
 
\end{tcolorbox}
\caption{Complete game trace for a successful game. The oracle feedback follows the $\to$ sign. The model correctly identifies the target in 7 turns. \textbf{T} = turn; \textbf{H} = current hypothesis; \textbf{R} = reasoning; \textit{positive} = positive test; \textit{negative} = negative test; \textit{guess} = guess turn.}
\label{tab:game1}
\end{table*}

\begin{table*}[ht]
\begin{tcolorbox}[
  colback=white,
  colframe=failcolor,
  coltitle=white,
  rounded corners,
  title={\sffamily Failed Game}
]
\scriptsize\setlength{\parskip}{1pt}

\textbf{target rule (hidden to player):} artifact \\
\textbf{Sampling Category (hidden to player):} electrical device \\
\textbf{Initial Items:} relay, primary coil, solenoid

\smallskip\hrule\smallskip
 
\textbf{T1} \textit{(positive)}\enspace
\textbf{Items:} transformer, inductor, electromagnet\enspace
\textbf{H:} Devices that operate using electromagnetic coils.\enspace
\textbf{R:} All initial examples are coil-based devices; testing others of the same kind.\enspace
$\to$ \texttt{CONFORM}
 
\smallskip\hrule\smallskip
 
\textbf{T2} \textit{(negative)}\enspace
\textbf{Items:} capacitor, resistor, diode\enspace
\textbf{H:} Electrical components relying on electromagnetic coils or inductive windings.\enspace
\textbf{R:} Testing components that do not use coils to check whether coil-dependency is required.\enspace
$\to$ \texttt{CONFORM}
 
\smallskip\hrule\smallskip
 
\textbf{T3} \textit{(negative)}\enspace
\textbf{Items:} gear, spring, pulley\enspace
\textbf{H:} Electrical or electronic components.\enspace
\textbf{R:} Testing purely mechanical components to probe whether the property is restricted to electrical items.\enspace
$\to$ \texttt{CONFORM}
 
\smallskip\hrule\smallskip
 
\textbf{T4} \textit{(negative)}\enspace
\textbf{Items:} happiness, freedom, velocity\enspace
\textbf{H:} Tangible physical objects (not abstract concepts).\enspace
\textbf{R:} Testing abstract concepts to verify whether tangibility is required.\enspace
$\to$ \texttt{DO NOT CONFORM}
 
\smallskip\hrule\smallskip
 
\textbf{T5} \textit{(guess)}\enspace
\textbf{Property:} ``tangible physical objects''\enspace
\textbf{R:} Abstract concepts do not conform; all physical objects tested so far do.\enspace
$\to$ \texttt{INCORRECT}
 
\smallskip\hrule\smallskip
 
\textbf{T6} \textit{(positive)}\enspace
\textbf{Items:} apple, mountain, table\enspace
\textbf{H:} The word has exactly two syllables.\enspace
\textbf{R:} Testing whether syllable count explains the conformance pattern.\enspace
$\to$ \texttt{DO NOT CONFORM}
 
\smallskip\hrule\smallskip
 
\textbf{T7} \textit{(positive)}\enspace
\textbf{Items:} engine, bolt, circuit\enspace
\textbf{H:} Technical or engineering-related terms.\enspace
\textbf{R:} All initial examples are engineering terms; testing whether this domain restriction defines the property.\enspace
$\to$ \texttt{CONFORM}
 
\smallskip\hrule\smallskip
 
\textbf{T8} \textit{(guess)}\enspace
\textbf{Property:} ``engineering-related terms''\enspace
\textbf{R:} All conforming items are engineering or technical terms.\enspace
$\to$ \texttt{INCORRECT}
 
\smallskip\hrule\smallskip
 
\textbf{T9} \textit{(positive)}\enspace
\textbf{Items:} keyboard, wheel, processor\enspace
\textbf{H:} Components or parts of a larger machine or system.\enspace
\textbf{R:} All examples so far are parts of larger systems; testing other machine components.\enspace
$\to$ \texttt{CONFORM}
 
\smallskip\hrule\smallskip
 
\textbf{T10} \textit{(negative)}\enspace
\textbf{Items:} rock, tree, river\enspace
\textbf{H:} Man-made components or parts of engineered systems.\enspace
\textbf{R:} Testing natural, non-manufactured objects to check whether being man-made is required.\enspace
$\to$ \texttt{DO NOT CONFORM}
 
\smallskip\hrule\smallskip
 
\textbf{T11} \textit{(guess)}\enspace
\textbf{Property:} ``man-made components of engineered systems''\enspace
\textbf{R:} Natural objects do not conform; all man-made items tested so far do.\enspace
$\to$ \texttt{INCORRECT}
 
\smallskip\hrule\smallskip
 
\textbf{T12} \textit{(negative)}\enspace
\textbf{Items:} chair, car, computer\enspace
\textbf{H:} The word contains the letter `o'.\enspace
\textbf{R:} Testing a letter-based pattern noticed across many conforming words.\enspace
$\to$ \texttt{CONFORM}
 
\smallskip\hrule\smallskip
 
\textbf{T13} \textit{(guess)}\enspace
\textbf{Property:} ``contains the letter `o'\,''\enspace
\textbf{R:} Chair, car, computer all contain `o' and conform.\enspace
$\to$ \texttt{INCORRECT}

\smallskip\hrule\smallskip

\textbf{T14} \textit{(negative)}\enspace
\textbf{Items:} desk, sofa, lamp\enspace
\textbf{H:} A machine or mechanical/electrical device (including components).\enspace
\textbf{R:} Testing man-made everyday objects that are not machines, to probe whether device-hood is required.\enspace $\to$ \texttt{CONFORM}

\smallskip\hrule\smallskip

\textbf{T15} \textit{(negative)}\enspace
\textbf{Items:} cat, dog, bird\enspace
\textbf{H:} Inanimate (non-living) things.\enspace
\textbf{R:} Testing living organisms to determine whether animacy is the distinguishing factor.\enspace
$\to$ \texttt{DO NOT CONFORM}
 
\smallskip\hrule\smallskip
 
\textbf{T16} \textit{(guess)}\enspace
\textbf{Property:} ``non-living things''\enspace
\textbf{R:} Living organisms do not conform; all non-living items tested so far do.\enspace
$\to$ \texttt{INCORRECT}
 
\smallskip\hrule\smallskip
 
\textbf{T17} \textit{(negative)}\enspace
\textbf{Items:} sky, ocean, stone\enspace
\textbf{H:} The word has exactly one syllable.\enspace
\textbf{R:} Re-examining whether syllable count distinguishes conforming from non-conforming items.\enspace
$\to$ \texttt{DO NOT CONFORM}
 
\smallskip\hrule\smallskip
 
\textbf{T18} \textit{(negative)}\enspace
\textbf{Items:} bus, train, plane\enspace
\textbf{H:} The word ends with a consonant.\enspace
\textbf{R:} Testing a phonological pattern as a potential systematic rule.\enspace
$\to$ \texttt{CONFORM}
 
\smallskip\hrule\smallskip
 
\textbf{T19} \textit{(guess)}\enspace
\textbf{Property:} ``countable nouns''\enspace
\textbf{R:} All conforming items are countable nouns; non-conforming items include abstractions and natural phenomena.\enspace
$\to$ \texttt{INCORRECT}
 
\smallskip\hrule\smallskip
 
\textbf{T20} \textit{(positive)}\enspace
\textbf{Items:} oxygen, gold, water\enspace
\textbf{H:} The word contains the letter `e'.\enspace
\textbf{R:} A final letter-based attempt, noting that many conforming words contain `e'.\enspace
$\to$ \texttt{DO NOT CONFORM}
 
\smallskip\hrule\smallskip
\textcolor{failcolor}{\textbf{Turn limit reached -- FAILED $\times$}}
 
\end{tcolorbox}
\caption{Complete game trace for a failed game. The oracle feedback follows the $\to$ sign. The game ends after the model used the 20-turn limit without identifying the target. Abbreviations as in Table~\ref{tab:game1}.}
\label{tab:game2}
\end{table*}
 
\end{document}